%% file: acl_latex.tex
\pdfoutput=1

\documentclass[11pt]{article}

\usepackage[preprint]{acl}

\usepackage{times}
\usepackage{latexsym}
\usepackage[T1]{fontenc}
\usepackage[utf8]{inputenc}
\usepackage{inconsolata}

\usepackage{microtype}
\usepackage{parskip}
\usepackage{lscape}

\usepackage{amsmath}
\usepackage{amssymb}
\usepackage{amsfonts}
\usepackage{amsthm}
\usepackage{nicefrac}

\usepackage{booktabs}
\usepackage{multirow}
\usepackage{colortbl}
\usepackage{threeparttable}

\usepackage{graphicx}
\usepackage{float}
\usepackage{placeins}
\usepackage{subcaption}

\usepackage[dvipsnames]{xcolor}
\definecolor{mygray}{gray}{.88}
\definecolor{pyblue}{rgb}{0.0, 0.0, 0.5}
\definecolor{pygreen}{rgb}{0.0, 0.5, 0.0}
\definecolor{pyorange}{rgb}{1.0, 0.4, 0.0}
\definecolor{myblue}{HTML}{5f8fb7}
\definecolor{myred}{HTML}{cc7c7c}
\definecolor{mybrown}{HTML}{D8C6A6}
\definecolor{mossgreen}{HTML}{8B9A7B}

\usepackage{enumitem}

\usepackage{algorithm}
\usepackage{algpseudocode}
\usepackage{listings}
\usepackage{alltt}

\lstdefinestyle{pythonstyle}{
    language=Python,
    basicstyle=\footnotesize\ttfamily,
    breaklines=true,
    morekeywords={self},
    keywordstyle=\color{pyblue},
    commentstyle=\color{pygreen},
    stringstyle=\color{pyorange},
    numberstyle=\tiny\color{gray},
    numbers=left,
    numbersep=10pt,
    tabsize=4,
    showspaces=false,
    showstringspaces=false
}

\usepackage[most]{tcolorbox}
\tcbuselibrary{breakable}

\tcbuselibrary{skins,breakable}
\usepackage{seqsplit}

\newcommand{\smiles}[1]{{\ttfamily\footnotesize\seqsplit{#1}}}

\newcommand{\skillcard}[6]{%
\begin{tcolorbox}[
  enhanced,
  colback=gray!4,
  colframe=black!25,
  boxrule=0.5pt,
  arc=1.5mm,
  left=1.2mm,right=1.2mm,top=0.8mm,bottom=0.8mm,
  title=\textbf{#1},
  fonttitle=\small,
]
{\small \textbf{Skill:} #2}\par\vspace{1mm}
{\scriptsize \textbf{Before:}}\par
\smiles{#3}\par\vspace{0.8mm}
{\scriptsize \textbf{After:}}\par
\smiles{#4}\par\vspace{0.8mm}
{\scriptsize \textbf{Score:} #5}\par
{\scriptsize \textbf{Note:} #6}
\end{tcolorbox}%
}

\usepackage{pifont}

\usepackage{tikz}

\usepackage{url}
\usepackage{cleveref}

\usepackage{xspace}
\usepackage{fancyvrb}
\usepackage{fvextra} 
\usepackage{titletoc}

\input{define.tex}

\input{math_commands}

\newlength\savewidth

\newcommand{\method}{{\fontfamily{lmtt}\selectfont \textbf{MolMem}}\xspace}

\title{\method: Memory-Augmented Agentic Reinforcement Learning for Sample-Efficient Molecular Optimization}

\author{
  \textbf{Ziqing Wang\textsuperscript{1}} \quad
  \textbf{Yibo Wen\textsuperscript{1}} \quad
  \textbf{Abhishek Pandey\textsuperscript{2}} \quad
  \textbf{Han Liu\textsuperscript{1}} \quad
  \textbf{Kaize Ding\textsuperscript{1}\thanks{Corresponding Author}} \\
  \textsuperscript{1}Northwestern University \quad
  \textsuperscript{2}AbbVie \\
  \texttt{\{ziqingwang2029, yibowen2024\}@u.northwestern.edu} \\
  \texttt{abhishek.pandey@abbvie.com} \\
  \texttt{\{hanliu, kaize.ding\}@northwestern.edu}
}

\begin{document}
\maketitle
\input{sec/0_abstract}

\input{sec/1_intro}

\input{sec/2_related_work}

\input{sec/3_preliminaries}
\input{sec/4_method}

\input{sec/5_experiments}
\input{sec/6_conclusion}

\input{sec/7_limitations}
\bibliography{main}

\appendix
\clearpage
\newpage

\section*{Appendix Table of Contents}
\addcontentsline{toc}{section}{Appendix Table of Contents}

\startcontents

{\small
\printcontents{}{1}{\setcounter{tocdepth}{2}}
}

\vspace{6pt}
\hrule
\vspace{10pt}

\input{sec/X_suppl}

\end{document}

%% file: define.tex
\definecolor{lightgray}{gray}{0.9} 
\definecolor{lightergray}{gray}{0.95} 



%% file: math_commands.tex
\usepackage{dsfont} 

\usepackage{amsmath,amsfonts,bm}

\def\eqref#1{equation~\ref{#1}}

\def\1{\bm{1}}

\DeclareMathAlphabet{\mathsfit}{\encodingdefault}{\sfdefault}{m}{sl}
\SetMathAlphabet{\mathsfit}{bold}{\encodingdefault}{\sfdefault}{bx}{n}



%% file: sec/0_abstract.tex
\begin{abstract}
In drug discovery, molecular optimization aims to iteratively refine a lead compound to improve molecular properties while preserving structural similarity to the original molecule. However, each oracle evaluation is expensive, making sample efficiency a key challenge for existing methods under a limited oracle budget. Trial-and-error approaches require many oracle calls, while methods that leverage external knowledge tend to reuse familiar templates and struggle on challenging objectives. A key missing piece is long-term memory that can ground decisions and provide reusable insights for future optimizations. To address this, we present \method{} (\textbf{Mol}ecular optimization with \textbf{Mem}ory), a multi-turn agentic reinforcement learning (RL) framework with a dual-memory system. Specifically, \method{} uses Static Exemplar Memory to retrieve relevant exemplars for cold-start grounding, and Evolving Skill Memory to distill successful trajectories into reusable strategies. Built on this memory-augmented formulation, we train the policy with dense step-wise rewards, turning costly rollouts into long-term knowledge that improves future optimization. Extensive experiments show that \method{} achieves 90\% success on single-property tasks (1.5$\times$ over the best baseline) and 52\% on multi-property tasks using only 500 oracle calls. Our code is available at \url{https://github.com/REAL-Lab-NU/MolMem}.
\end{abstract}

%% file: sec/1_intro.tex
\section{Introduction}
\label{sec:intro}

Molecular optimization is a key step in drug discovery that iteratively refines a lead compound to improve molecular properties while preserving structural similarity to the original molecule~\citep{plowright2012hypothesis, wesolowski2016strategies, wang2025survey, wang2025polo}. Since each refinement relies on expensive oracle evaluations (e.g., wet-lab assays, high-fidelity simulators, or property predictors), \textbf{sample efficiency} becomes the core challenge~\citep{gao2022sample, guo2024augmented}: how to achieve strong optimization performance under a limited oracle budget?

\begin{figure}[t]
   \centering
   \includegraphics[width=\linewidth]{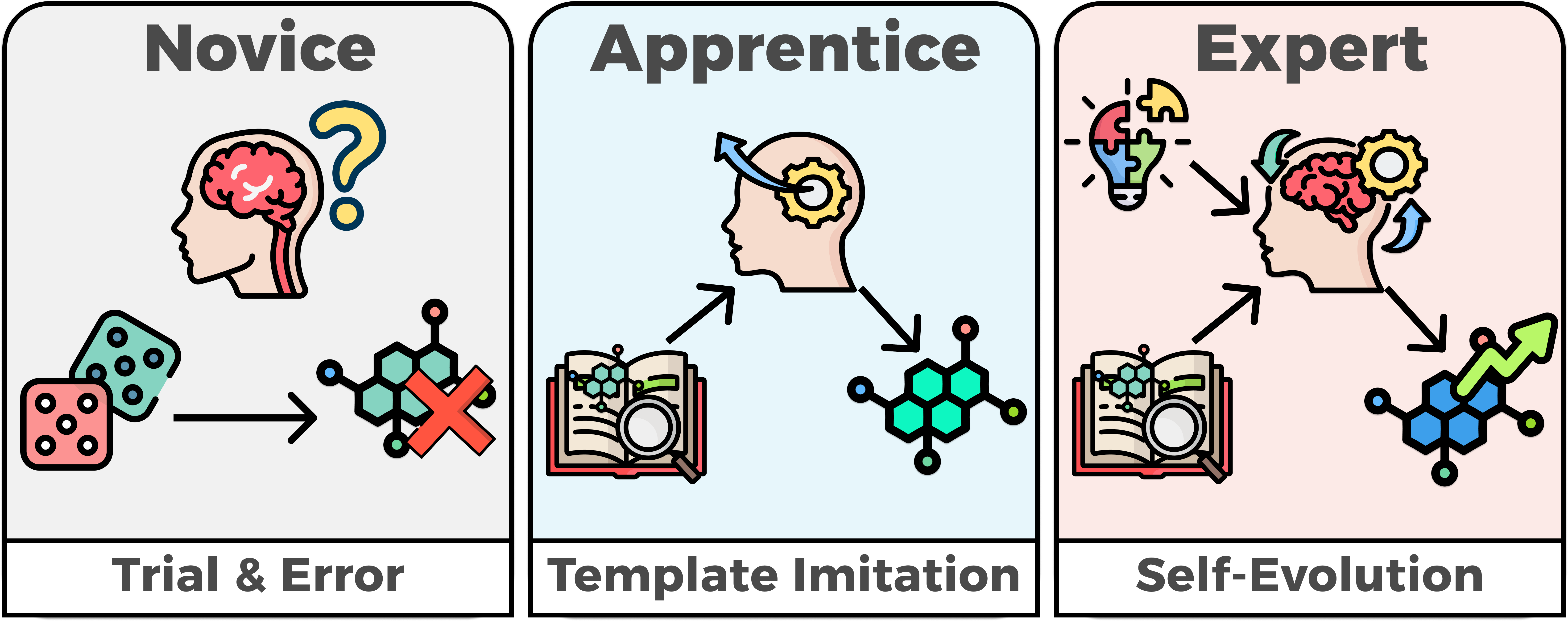}
   \vspace{-6mm}
\caption{\small\textbf{The Evolution of Molecular Optimization Paradigms.} 
(1) \textbf{Novice}: trial-and-error exploration with low sample efficiency; 
(2) \textbf{Apprentice}: template imitation of external knowledge; 
(3) \textbf{Expert (Ours)}: self-evolution via grounded knowledge and consolidated skills.}
    \label{fig:evolution}
   \vspace{-5mm}
\end{figure}

\vspace{-1mm}
However, most existing methods still struggle with this sample-efficiency challenge. To understand why, we highlight two common paradigms in Fig.~\ref{fig:evolution} and contrast them with how human experts work. \ding{182}\ \textbf{Novice} methods rely on trial-and-error search driven by oracle feedback, such as genetic algorithms and single-step reinforcement learning~\citep{jensen2019graph, olivecrona2017molecular, loeffler2024reinvent}. Without prior knowledge, they typically require many oracle calls to reach good performance. \ding{183}\ \textbf{Apprentice} methods leverage external knowledge such as large offline datasets, pretrained models, or retrieved exemplars~\citep{guo2023can, liu2024conversational, ye2025drugassist}. They often start faster by imitating familiar transformation templates, but struggle when the task demands edits beyond those templates. \ding{184}\  Human experts, in contrast, ground decisions in relevant references and turn successful trials into reusable strategies. This ability to learn from experience is key to sample-efficient optimization.

\vspace{-1mm}
Agentic reinforcement learning (RL) provides a natural framework for capturing this ability: an agent interacts with an environment over multiple turns and learns from feedback~\citep{wang2025ragen, zhang2025landscape}. Using large language models (LLMs) as the agent backbone makes this framework practical in this domain, since they can follow instructions and incorporate rich textual context to guide molecular edits~\citep{wang2025survey}. However, this multi-turn agentic framework remains under-explored in molecular optimization. More importantly, most existing methods provide limited support for long-term memory: once a rollout ends, useful discoveries are lost and cannot be reused in future optimizations. Such memory is crucial for sample efficiency, since discoveries from one run can reduce oracle calls in the next. This leads to a key question:

\begin{tcolorbox}[
    enhanced,
    sidebyside, 
    colframe=black!70,
    colback=yellow!5,
    boxrule=1pt, 
    arc=3mm,
    lefthand width=0.08\linewidth,
    sidebyside gap=3mm,
    left=0mm,          
    right=1mm,          
    top=1mm,          
    bottom=1mm,       
    before skip=5pt,    
    after skip=15pt,    
    sidebyside align=center,  
]
\includegraphics[width=1.1\linewidth]{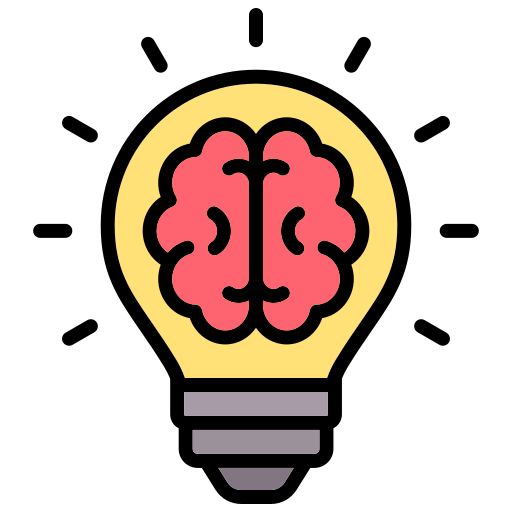}
\tcblower
\small\textit{Can we distill multi-turn rollouts into long-term, retrievable skills for sample-efficient optimization?}
\end{tcolorbox}

\begin{figure*}[t]
   \includegraphics[width=\linewidth]{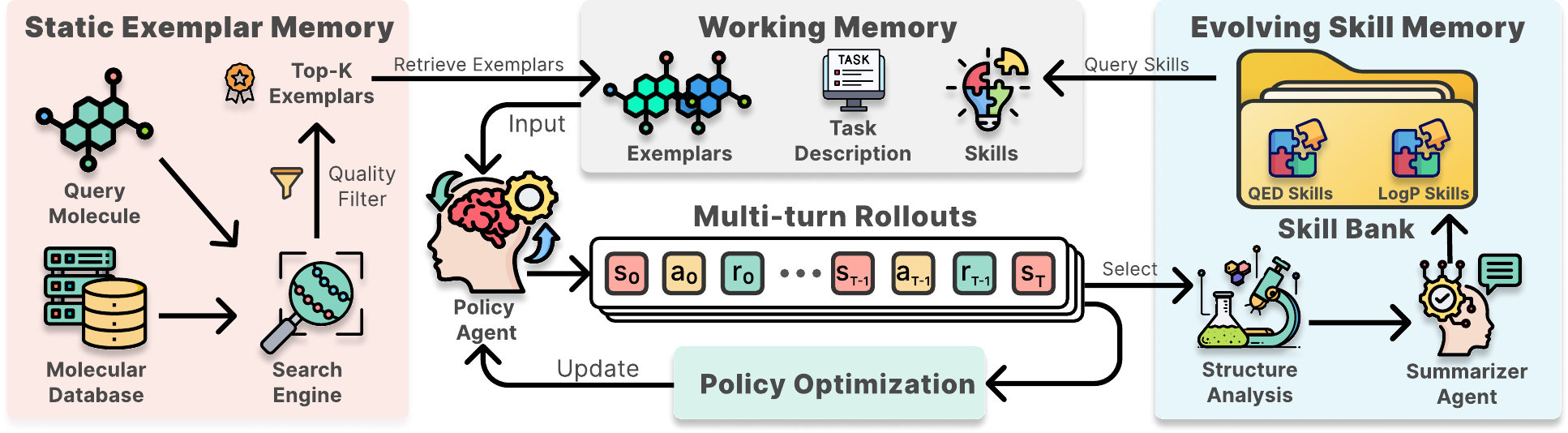}
   \vspace{-6mm}
\caption{\small\textbf{Overview of the \method{}.}
The framework consists of two complementary memory modules:
\textbf{(Left) Static Exemplar Memory} retrieves structurally similar, high-scoring molecules from an external database as references for the current objective.
\textbf{(Right) Evolving Skill Memory} distills successful trajectories into reusable textual strategies.
Retrieved exemplars or skills are injected into the \textbf{Working Memory} (center) to augment the LLM agent's context during \textbf{Multi-turn Rollouts}.}

   \label{fig:overview}
   \vspace{-3mm}
\end{figure*}

We answer this question with \method{} (\textbf{Mol}ecular optimization with \textbf{Mem}ory), a multi-turn agentic RL framework with a dual-memory system (Fig.~\ref{fig:overview}). Within each rollout, the agent uses the trajectory history as short-term context. To complement this, \method{} maintains two long-term memory components: \ding{182}\ \textbf{Static Exemplar Memory} retrieves relevant exemplars (e.g., structurally similar high-scoring molecules) to provide cold-start grounding when the agent lacks its own experience; \ding{183}\ \textbf{Evolving Skill Memory} distills successful trajectories into reusable strategies stored in a skill bank, enabling the agent to improve beyond directly copying exemplars. Long-term memory is queried only when optimization plateaus, encouraging the agent to explore independently before consulting retrieved guidance. To train the multi-turn policy effectively, we use dense step-wise rewards derived from oracle signals, enabling precise credit assignment across the trajectory. To summarize, our main contributions are as follows:
\begin{itemize}[leftmargin=*,itemsep=-1pt]
    \item \textbf{Framework.} We propose \method{}, a memory-augmented multi-turn agentic RL framework for sample-efficient molecular optimization.
    
    \item \textbf{Memory System.} We design a dual-memory system with distinct roles: Static Exemplar Memory for cold-start grounding, and Evolving Skill Memory for distilling successful trajectories into reusable strategies.
    
    \item \textbf{Effectiveness.} Experiments show that \method{} achieves \textbf{90\%} success on single-property tasks  (\textbf{1.5$\times$} over best baseline) and \textbf{52\%} on multi-property tasks using only \textbf{500} oracle calls.
\end{itemize}

%% file: sec/2_related_work.tex
\vspace{-1mm}
\section{Related Work} \label{sec: related_work}

\paragraph{Molecular Optimization.}
To reduce the cost of wet-lab assays, researchers have proposed various computational methods for molecular optimization~\citep{gao2022sample}. Approaches include genetic algorithms~\citep{jensen2019graph}, Bayesian optimization~\citep{korovina2020chembo}, and reinforcement learning for goal-directed generation~\citep{popova2018deep, olivecrona2017molecular, loeffler2024reinvent}. However, the PMO benchmark shows that many of these methods struggle under realistic oracle budgets and often have difficulty balancing property improvement with structural similarity to the original lead~\citep{gao2022sample}. More recently, LLMs have been explored as flexible editors for interactive molecular design~\citep{ye2025drugassist, liu2024conversational, wang2024efficient, rivera2026glassmol}. Despite their flexibility, these methods are typically used in a single-step manner and do not systematically learn from full optimization trajectories, which limits cross-trajectory improvement under tight budgets. This motivates multi-turn formulations that optimize over trajectories with step-wise feedback.

\paragraph{Multi-Turn Agentic Reinforcement Learning.}
Multi-turn reinforcement learning provides a natural way to model sequential decision making with LLM agents~\citep{du2023improving, wang2024rethinking}. It also aligns well with molecular optimization, where an optimizer proposes a sequence of edits under step-wise oracle feedback. Under the broader umbrella of agentic RL, LLMs are treated as policies that learn through interaction with an environment rather than as one-shot generators~\citep{zhang2025landscape}. Recent systems such as RAGEN~\citep{wang2025ragen} show that optimizing over full trajectories can improve long-horizon behavior. While multi-turn RL has advanced rapidly in language-based tasks, its use for sample-efficient molecular optimization remains relatively under-explored. Moreover, most multi-turn setups primarily use the trajectory history as short-term context: once an episode ends, useful intermediate insights are difficult to carry over to future runs~\citep{wang2025polo}. This gap motivates mechanisms that can store and retrieve experience across trajectories.

\paragraph{Memory-Augmented Agents.}
A growing line of work equips LLM agents with memory to reuse information and experience across steps and episodes~\citep{hu2025memory}. In molecular design, memory is often implemented as static retrieval: systems such as ChatDrug~\citep{liu2024conversational} and DrugAssist~\citep{ye2025drugassist} retrieve similar molecules or references from databases to guide editing, but this context is typically used as read-only guidance rather than being updated from optimization outcomes. Beyond molecular design, general-purpose agents have explored experience-centric memory, where interaction feedback is consolidated into reusable artifacts. For example, Voyager builds a library of executable code skills from task feedback~\citep{wang2023voyager}, and ExpeL distills successful trajectories into exemplars for future reuse~\citep{zhao2024expel}. Chemistry-focused agents have also begun to build self-updating libraries for general chemistry tasks (e.g., ChemAgent)~\citep{tang2025chemagent}. However, these ideas have not been tailored to molecular optimization, where an agent must improve properties while preserving structural similarity under tight oracle budgets. \method{} builds on this direction by combining static exemplar retrieval for grounding with trajectory-based skill distillation for reuse across runs.

%% file: sec/3_preliminaries.tex
\section{Preliminary}
\label{sec:preliminary}

\subsection{Problem Formulation}
\label{sec:problem}

Molecular optimization aims to refine a lead compound by proposing structural edits that improve one or more molecular properties under a limited oracle-call budget. Let $\mathcal{M}_{\text{mol}}$ denote the space of valid molecules. Given a lead molecule $m \in \mathcal{M}_{\text{mol}}$, the goal is to find an optimized molecule $m' \in \mathcal{M}_{\text{mol}}$ that solves:
\begin{equation}
\resizebox{\columnwidth}{!}{$
\max_{m' \in \mathcal{M}_{\text{mol}}} \sum_{i=1}^{n} w_i F_i(m')
\;\text{s.t.}\;
\begin{cases}
\mathrm{sim}(m,m') \ge \gamma,\\
\sum_{i=1}^{n} c_i \le B,
\end{cases}
$}
\end{equation}
where $F_i: \mathcal{M}_{\text{mol}} \rightarrow \mathbb{R}$ are black-box property oracles (e.g., binding affinity, solubility), $w_i$ are their weights, $\mathrm{sim}(\cdot,\cdot)$ denotes Tanimoto similarity with threshold $\gamma$ to encourage structural similarity, $c_i$ denotes the oracle call of evaluating $F_i$, and $B$ is the total oracle-call budget.

\subsection{Multi-Turn MDP Formulation}
\label{sec:mdp}
We model molecular optimization as a finite-horizon Markov Decision Process (MDP) in which an agent iteratively proposes candidate molecules and receives oracle evaluations after each turn. Formally, the MDP is defined as $\langle \mathcal{S}, \mathcal{A}, P, R \rangle$ with horizon $T$:
\begin{itemize}[leftmargin=*,topsep=0pt]
   \item \textbf{State $\mathcal{S}$.} At turn $t$, the state $s_t$ contains the task objective and the optimization context, including the lead molecule $m_0$, the previously proposed molecules $(m_1,\ldots,m_t)$, and their reward $(r_0,\ldots,r_{t-1})$.
   \item \textbf{Action $\mathcal{A}$.} The action $a_t \sim \pi_\theta(\cdot \mid s_t)$ corresponds to proposing the next candidate molecule $m_{t+1}$, represented as a SMILES string.
   \item \textbf{Transition and reward $P, R$.} After taking action $a_t$, the environment evaluates the proposed molecule $m_{t+1}$ with oracle(s) and returns a step reward $r_t = R(s_t,a_t)$ (Appendix~\ref{appendix:reward}). The next state $s_{t+1}$ is obtained by updating the history with $(m_{t+1}, r_t)$. The transition is written as $s_{t+1} \sim P(\cdot \mid s_t, a_t)$.
\end{itemize}
This multi-turn formulation allows the agent to learn from intermediate feedback over a trajectory, rather than optimizing each edit in isolation.

%% file: sec/4_method.tex
\section{Method}
\label{sec:method}

We present \method{}, a multi-turn agentic reinforcement learning (RL) approach for sample-efficient molecular optimization. Building on the MDP formulation in Section~\ref{sec:mdp}, we first describe the dual-memory system (Section~\ref{sec:memory}) and then introduce the policy optimization procedure. (Section~\ref{sec:policy_opt}).

\subsection{Agentic Memory System}
\label{sec:memory}

In the multi-turn MDP, the trajectory history provides short-term context within a rollout. However, once a rollout ends, the agent has no mechanism to retain and reuse effective edits discovered along the way, so similar oracle calls may be repeated across rollouts. To support reuse for better sample efficiency, \method{} maintains a dual-memory system $\mathcal{M} = (\mathcal{M}^{\text{static}}, \mathcal{M}^{\text{evolve}})$.

\subsubsection{Static Exemplar Memory}
The static component $\mathcal{M}^{\text{static}}$ provides \textbf{cold-start grounding} by maintaining a large molecule bank constructed from ChEMBL~\citep{zdrazil2024chembl} (2.8M molecules) with precomputed physicochemical properties (e.g., QED, LogP). Each molecule is indexed by its Morgan fingerprint (ECFP4; radius $=2$, 2048-bit) using FAISS~\citep{johnson2019billion} for efficient similarity search.

To avoid over-reliance on external guidance, we trigger exemplar retrieval only when progress plateaus (e.g., no reward improvement for consecutive turns), rather than at every turn or from the start. This encourages independent exploration in early turns before incorporating exemplar-based grounding. Given the current molecule $m_t$ at the triggered turn and the lead molecule $m_0$, retrieval follows a two-stage procedure:

\paragraph{Candidate Retrieval.} We first  perform approximate nearest-neighbor retrieval in fingerprint space to obtain a candidate set:
\begin{equation}
\label{eq:ex_coarse}
\mathcal{C}_t = \textsc{ANN}\!\left(\phi(m_t);\,\mathcal{M}^{\text{static}}\right),
\end{equation}
where $\phi(\cdot)$ denotes the ECFP4 fingerprint and \textsc{ANN} is implemented with FAISS.

\paragraph{Constrained Reranking.} We then enforce lead similarity by filtering candidates with Tanimoto similarity to $m_0$, and return the top-$K$ exemplars ranked by target property score:
\begin{equation}
\label{eq:ex_fine_topk}
c_t^{\mathrm{static}}
=\mathrm{Top}\text{-}K_{\,m'\in\mathcal{C}_t:\ \mathrm{sim}(m',m_0)\ge \gamma_{\mathrm{ex}}}\;
F_{\mathcal Q}(m').
\end{equation}
where $\mathrm{sim}(\cdot,\cdot)$ is Tanimoto similarity, $\gamma_{\mathrm{ex}}$ is a similarity threshold, and $F_{\mathcal{Q}}(\cdot)$ is the task-specific score. This design uses $m_t$ for broad retrieval while enforcing similarity to $m_0$, ensuring retrieved exemplars both explore relevant chemical space and remain similar to the lead.

Crucially, exemplars serve as \textit{references} rather than targets to copy. We penalize exact copying by assigning a negative reward if the generated molecule is identical to any retrieved exemplar, discouraging rote replication and encouraging the agent to learn from structural patterns. Implementation details are provided in Appendix~\ref{appendix:retrieval}.

\begin{figure}[t]
   \centering
   \includegraphics[width=\linewidth]{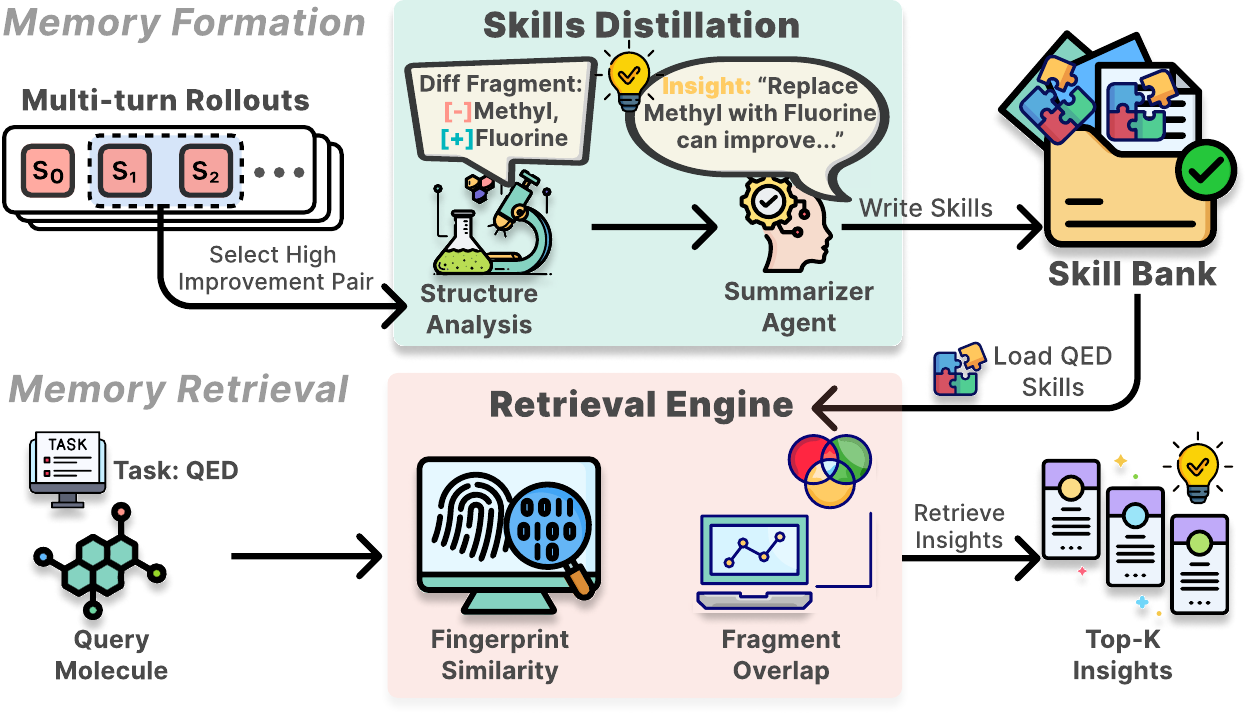}
   \vspace{-6mm}
    \caption{\small\textbf{Mechanism of the Evolving Skill Memory.} 
     \textbf{(Top) Memory Formation}: High-reward pairs in multi-turn rollouts are distilled into textual skills by extracting structural changes and synthesizing insights via a summarizer agent.
     \textbf{(Bottom) Memory Retrieval}: Relevant skills are retrieved from the skill bank using a hybrid matching of fingerprint similarity and functional group overlap to guide the policy agent.}
    \label{fig:skill_memory}
   \vspace{-2mm}
\end{figure}

\subsubsection{Evolving Skill Memory}
Unlike the static, external exemplar bank, the evolving component $\mathcal{M}^{\text{evolve}}$ grows a library of reusable strategies consolidated from the agent's own high-reward rollouts (Fig.~\ref{fig:skill_memory}). We implement this component via \textit{memory formation} and \textit{memory retrieval}.

\paragraph{Memory Formation.}
After each training iteration, we extract step-wise experiences from multi-turn rollouts.
For each transition $(m_t, m_{t+1})$ with high reward improvement $\Delta r  > \delta$ (where $\delta$ is a threshold), 
we build a structured \emph{edit card} $\kappa_t$ that summarizes the transformation, including:
(i) MCS-based edit decomposition (modification type and removed/added fragments),
(ii) scaffold analysis (before/after scaffold and scaffold type, e.g., scaffold hop),
(iii) functional-group additions/removals, and
(iv) cheap descriptor deltas (e.g., MW, PSA, HBD/HBA, ring count).
A summarizer LLM then converts this card into a single actionable strategy sentence following a fixed \emph{Action--What--Where--Effect} template, e.g., \emph{``Replace methoxy (-OCH$_3$) with fluorine (-F) on the aromatic ring to improve the target score.''}
\begin{equation}
\label{eq:skill_summarize}
e = \mathrm{LLM}_{\mathrm{summarizer}}\!\left(\kappa_t,\,\Delta r,\,\mathcal Q\right).
\end{equation}
Each skill $e$ is stored in a task-specific bank $\mathcal{M}^{\text{evolve}}_{\mathcal Q}$ and indexed by the molecule $m_t$'s Morgan fingerprint and a set of functional-group tags.

\paragraph{Memory Retrieval.}
Given the current molecule $m_t$ and objective $\mathcal Q$, we query $\mathcal{M}^{\text{evolve}}_{\mathcal Q}$ via two parallel matching signals. Fingerprint matching uses Tanimoto similarity over Morgan fingerprints, and functional-group matching uses Jaccard similarity over functional-group sets. Each skill $e$ is stored with its pre-edit source molecule $m(e)$ and functional-group tags $G(e)$; we compute $\mathrm{sim}_{\mathrm{FP}}(m_t,e)=\mathrm{sim}(\phi(m_t),\phi(m(e)))$ and $\mathrm{sim}_{\mathrm{FG}}(m_t,e)=J(G(m_t),G(e))$. After threshold filtering, we return top-$K_{\text{fp}}$ and top-$K_{\text{fg}}$ skills respectively:
\begin{equation}
\label{eq:skill_retrieve}
\scalebox{0.90}{$
\begin{aligned}
c_t^{\mathrm{fp}} &=
\mathrm{Top}\text{-}K_{\text{fp}}\!\left\{
e\in\mathcal{M}^{\text{evolve}}_{\mathcal Q}:\,
\mathrm{sim}_{\mathrm{FP}}(m_t,e)\ge\gamma_{\mathrm{fp}}
\right\},\\
c_t^{\mathrm{fg}} &=
\mathrm{Top}\text{-}K_{\text{fg}}\!\left\{
e\in\mathcal{M}^{\text{evolve}}_{\mathcal Q}:\,
\mathrm{sim}_{\mathrm{FG}}(m_t,e)\ge\gamma_{\mathrm{fg}}
\right\},
\end{aligned}
$}
\end{equation}
and set $c_t^{\mathrm{sk}} = c_t^{\mathrm{fp}} \cup c_t^{\mathrm{fg}}$. Among candidates that pass the threshold, we rank skills by their improvement $\Delta r$, ensuring retrieved skills are not only structurally relevant but also empirically effective.

Similar to exemplar retrieval, skill retrieval is triggered only when optimization plateaus, encouraging the agent to first explore independently before consulting accumulated experience. Retrieved skills are injected into the agent's working memory as high-level guidance, enabling cross-rollout reuse of successful edit principles without additional oracle evaluations. Implementation details are in Appendix~\ref{appendix:skill}.

\subsubsection{Memory-Augmented Rollouts}
As described above, exemplar and skill retrieval are triggered only when optimization plateaus. If both trigger conditions are satisfied simultaneously, we stochastically select one memory source to keep the LLM context budget and prevent over-reliance on a single modality. 
The retrieved items are assembled into the policy input (working memory), forming the memory-augmented state:
\begin{equation}
\label{eq:state}
s_t = \big(\mathcal{Q},\; \mathcal{H}_t \oplus c_t^{\mathrm{mem}}\big),
\end{equation}
where $\mathcal{Q}$ is the task objective, $\mathcal{H}_t = (m_0, m_1, \ldots, m_t; r_0, \ldots, r_{t-1})$ is the trajectory history, $\oplus$ denotes context concatenation, and $c_t^{\mathrm{mem}} \in \{c_t^{\mathrm{static}}, c_t^{\mathrm{evolve}}, \varnothing\}$ denotes the injected memory depending on trigger status. This design combines within-rollout context with cross-rollout knowledge via on-demand retrieval.

\subsection{Memory-Augmented Policy Optimization}
\label{sec:policy_opt}

Given the memory-augmented state formulation (Eq.~\ref{eq:state}), we train \method{} in two stages:
\paragraph{Stage I: Supervised Fine-Tuning.}
We initialize the policy $\pi_\theta$ via supervised fine-tuning (SFT) on an offline dataset of molecular edit pairs that satisfy the similarity constraint. This warm start teaches the model to propose valid edits and provides a stable prior before RL (details in Appendix~\ref{appendix:sft}).

\paragraph{Stage II: Memory-Aware Policy Optimization.}
Building on the SFT-initialized policy, we apply Proximal Policy Optimization (PPO)~\citep{schulman2017proximal} to further refine the agent's multi-turn decision making. Unlike single-turn methods that treat each edit independently, our formulation optimizes over entire trajectories while providing \emph{dense step-wise feedback}: the agent receives an immediate reward $r_t$ after each modification, enabling precise credit assignment rather than relying solely on terminal outcomes. Given a rollout $\tau$ collected by the policy $\pi_{\theta_{\text{old}}}$, we optimize:
\begin{equation}
\begin{aligned}
J_{\text{PPO}}(\theta)
&=
\mathbb{E}_{\tau \sim \pi_{\theta_{\text{old}}}}
\left[
\sum_{t=0}^{T-1} L_t(\theta)
\right], \\
L_t(\theta)
&=
\min\Big(
\rho_t \hat{A}_t,\;
\mathrm{clip}(\rho_t, 1{-}\epsilon, 1{+}\epsilon)\hat{A}_t
\Big),
\end{aligned}
\label{eq:ppo_step}
\end{equation}
where $\rho_t = \frac{\pi_\theta(a_t \mid s_t)}{\pi_{\theta_{\text{old}}}(a_t \mid s_t)}$ is the importance ratio and $\hat{A}_t$ is computed via GAE~\citep{schulman2015high} from the dense reward sequence $(r_t, r_{t+1}, \ldots)$, which balances property improvement with similarity preservation (details in Appendix~\ref{appendix:reward}). Since the state $s_t$ (Eq.~\ref{eq:state}) incorporates retrieved memory when triggered (Section~\ref{sec:memory}), training naturally teaches the policy to interpret and leverage memory guidance for long-term improvement.

%% file: sec/5_experiments.tex
\section{Experiments}  \label{sec: exp}

\input{tables/single_property}

\subsection{Experimental Setup}
\label{sec:expt:setup}

\textbf{Baselines.} We compare \method{} with baselines from two \emph{baseline} paradigms (Fig.~\ref{fig:evolution}):
\textbf{(1) Novice:} Graph-GA~\citep{jensen2019graph}, QMO~\citep{hoffman2022optimizing}, and Reinvent 4~\citep{loeffler2024reinvent};
\textbf{(2) Apprentice:}
(i) \textbf{Direct Retrieval}, a retrieval-only variant that retrieves molecules using the same static retriever as \method{};
(ii) \textbf{Direct Prompt} with general-purpose LLMs using the same instruction prompt as \method{} for fair comparison;
(iii) \textbf{SFT-only}, using the same supervised data and recipe as Stage~I but without RL;
(iv) \textbf{Task-Specific LLMs:} MOLLEO~\citep{wang2024efficient}, LlaSMol~\citep{yu2024llasmol}, ChemLLM~\citep{zhang2024chemllm}, PEIT-LLM~\citep{lin2024property}, and GeLLM$^3$O~\citep{dey2025mathtt}.
Direct Retrieval and SFT-only also serve as ablations of \method{}. 
Notably, \method{} uses a compact \textbf{Qwen2.5-1.5B} backbone, while most task-specific LLM baselines use 7--8B parameters.
Details of these baselines are in Appendix~\ref{appendix:baselines}.

\vspace{-2mm}
\textbf{Tasks and Constraints.} We evaluate on five single-property tasks (QED, plogP, SA, DRD2, JNK3) and five multi-property tasks (QED+plogP, plogP+DRD2, QED+SA, DRD2+SA, DRD2+QED+plogP) using 200 lead molecules randomly sampled from ZINC-250k~\citep{irwin2005zinc}. We enforce two practical constraints: (1) a Tanimoto similarity threshold of $\gamma=0.4$ to preserve similarity to the lead, and (2) an oracle-call budget of $B=500$ per lead molecule.

\vspace{-2mm}
\textbf{Evaluation Metrics.} Following prior work~\citep{dey2025mathtt}, we report three complementary metrics:
\textbf{(1) Success Rate (SR):} the percentage of leads that meet the task-specific success criterion while satisfying the similarity constraint;
\textbf{(2) Similarity (Sim):} the average Tanimoto similarity between the lead and the final optimized molecule;
\textbf{(3) Relative Improvement (RI):} the average relative improvement from the lead to the final molecule over the target property.
Detailed definitions and task-specific success criteria are provided in Appendix~\ref{appendix:metrics}.

\vspace{-2mm}
\textbf{Agent Engine.} We use GPT-4o as the skill summarizer in \method{} (Appendix~\ref{appendix:skill}).

\input{tables/multi_properties}

\begin{figure*}[t!]
\centering
\begin{subfigure}[b]{0.32\textwidth}
    \centering
    \includegraphics[width=\textwidth]{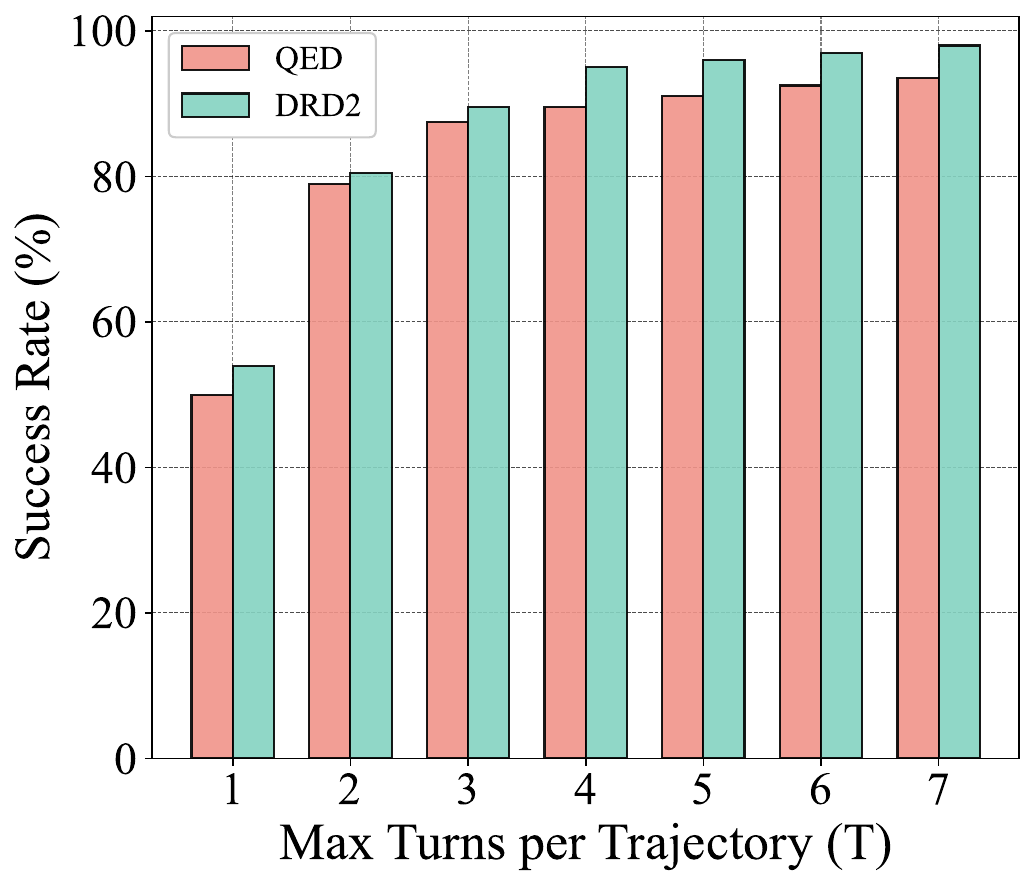}
    \caption{\small Impact of Trajectory Length ($T$)}
    \label{fig:traj_length}
\end{subfigure}
\hfill
\begin{subfigure}[b]{0.32\textwidth}
    \centering
    \includegraphics[width=\textwidth]{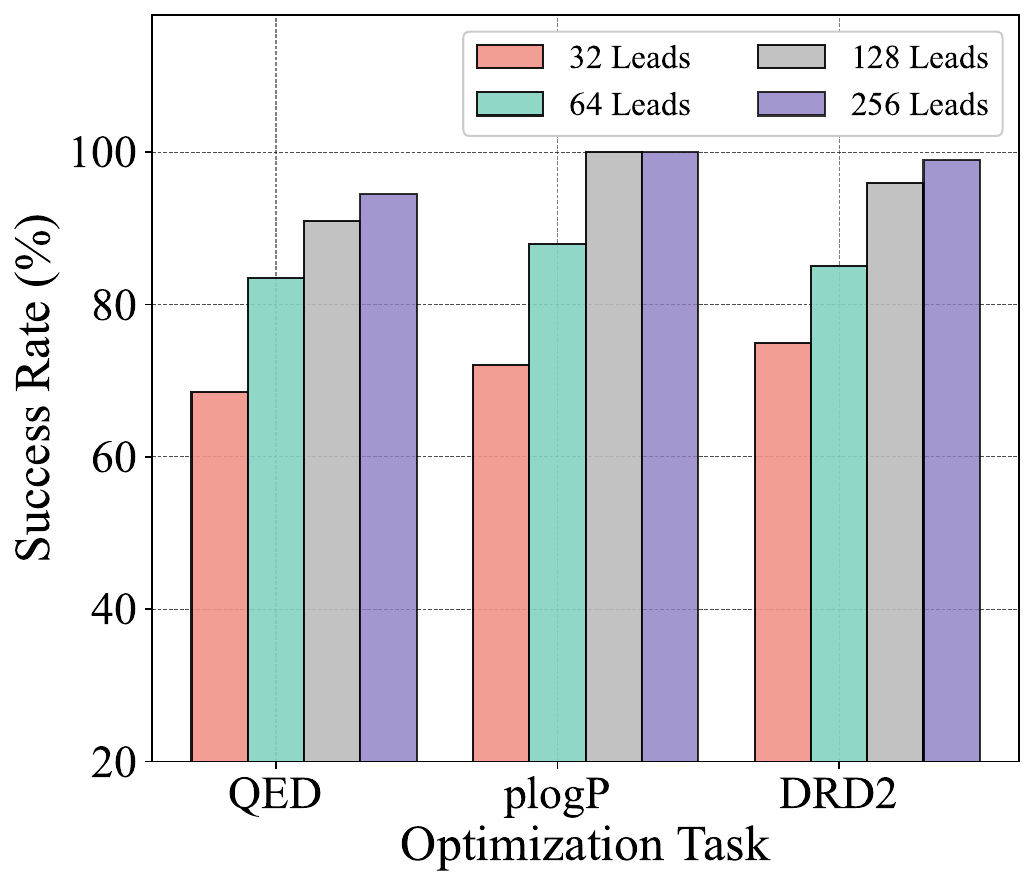}
    \caption{\small Impact of Training Data Size}
    \label{fig:train_size}
\end{subfigure}
\hfill
\begin{subfigure}[b]{0.32\textwidth}
    \centering
    \includegraphics[width=\textwidth]{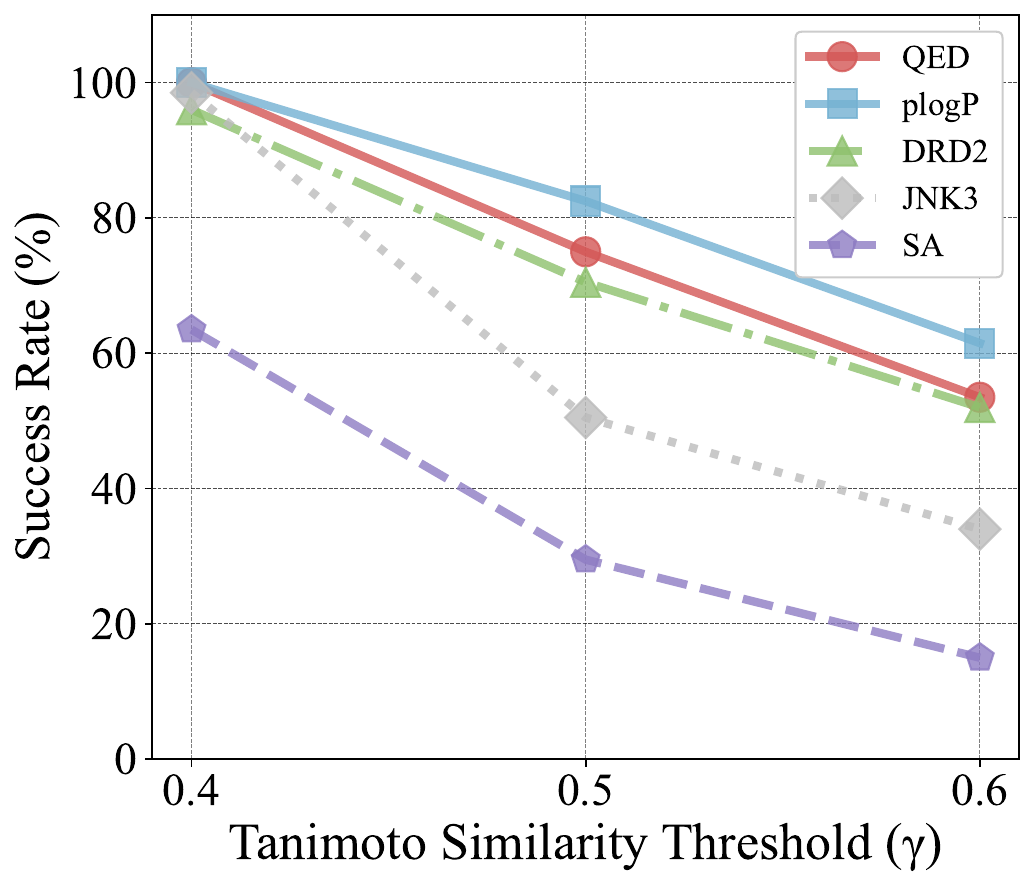}
    \caption{\small Impact of Similarity Constraint ($\gamma$)}
    \label{fig:sim_threshold}
\end{subfigure}
\vspace{-2mm}
\caption{\small\textbf{Analysis of key hyperparameters.} 
(a) Success rate improves with more turns and saturates around $T{=}5$. 
(b) Performance improves with more training leads, though even 64 leads yield competitive results.
(c) Stricter similarity constraints reduce the success rate.}
\label{fig:sensitivity}
\vspace{-3mm}
\end{figure*}

\subsection{Single-Property Optimization}
\label{sec:single_property}
Table~\ref{tab:single} reports five single-property tasks. \method{} achieves the highest success rate across all tasks while maintaining comparable similarity.
\vspace{-0.15cm}
\paragraph{Largest gains emerge on bioactivity targets.}
Compared with strong task-specific LLM baselines, \method{} improves SR from 82.5\% to 91.0\% on QED and from 81.5\% to 100.0\% on plogP. The gap widens on bioactivity targets: 50.5\% to 96.0\% on DRD2 and 44.0\% to 98.5\% on JNK3. These tasks require precise navigation of structure-activity relationships, suggesting that \method{} is more effective on such complex landscapes.
\vspace{-0.15cm}
\paragraph{Smaller backbone, larger gains.}
Despite using a compact Qwen2.5-1.5B backbone, \method{} outperforms much larger baselines (ChemLLM 7B, GeLLM$^3$O 8B, PEIT-LLM 8B). The multi-turn framework compensates for limited model capacity by allowing iterative refinement, while the dual-memory system supplies chemical knowledge to retrievable exemplars and accumulated skills, reducing the need to encode such knowledge purely in model parameters.
\vspace{-0.15cm}
\paragraph{Retrieval provides cold-start, learning enables scaling.}
Direct Retrieval performs well on QED (68.5\% SR) but struggles on bioactivity targets (28.5\% on DRD2 and 21.0\% on JNK3), where relevant exemplars are scarce. \method{} sustains improvement by learning from multi-turn RL and distilling successful edits into reusable skills, moving beyond exemplar copying.

\subsection{Multi-Property Optimization}
\label{sec:multi_property}

Table~\ref{tab:multi} reports five multi-property tasks. \method{} achieves the highest success rate on all tasks, with particularly large margins on combinations involving bioactivity targets. For example, on DRD2+SA, \method{} reaches 69.5\% SR compared to 37.5\% for a strong baseline (Reinvent 4). The three-property task (DRD2+QED+plogP) proves challenging for all methods, yet \method{} still leads with 15.0\% versus 8.0\% (Direct Retrieval). These results suggest that the benefits of memory-augmented multi-turn optimization extend from single-property to multi-property settings, especially when the objective involves complex bioactivity constraints.

\begin{figure*}[t]
\centering
\begin{subfigure}[b]{0.32\textwidth}
    \centering
    \includegraphics[height=3.3cm]{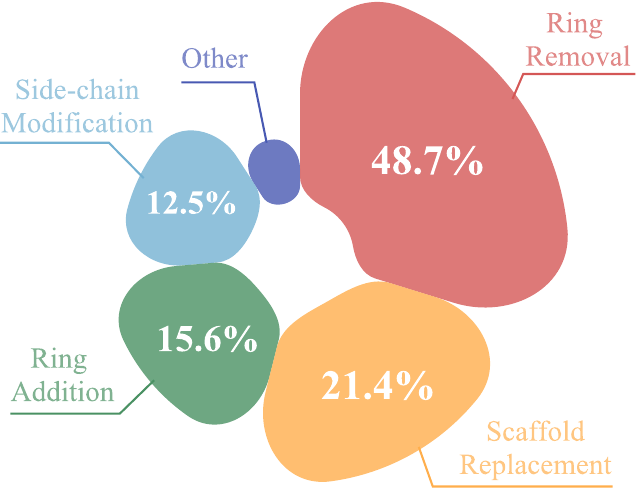}
    \caption{Scaffold Modification Types}
    \label{fig:scaffold}
\end{subfigure}
\hfill
\begin{subfigure}[b]{0.33\textwidth} 
    \centering
    \includegraphics[height=3.3cm]{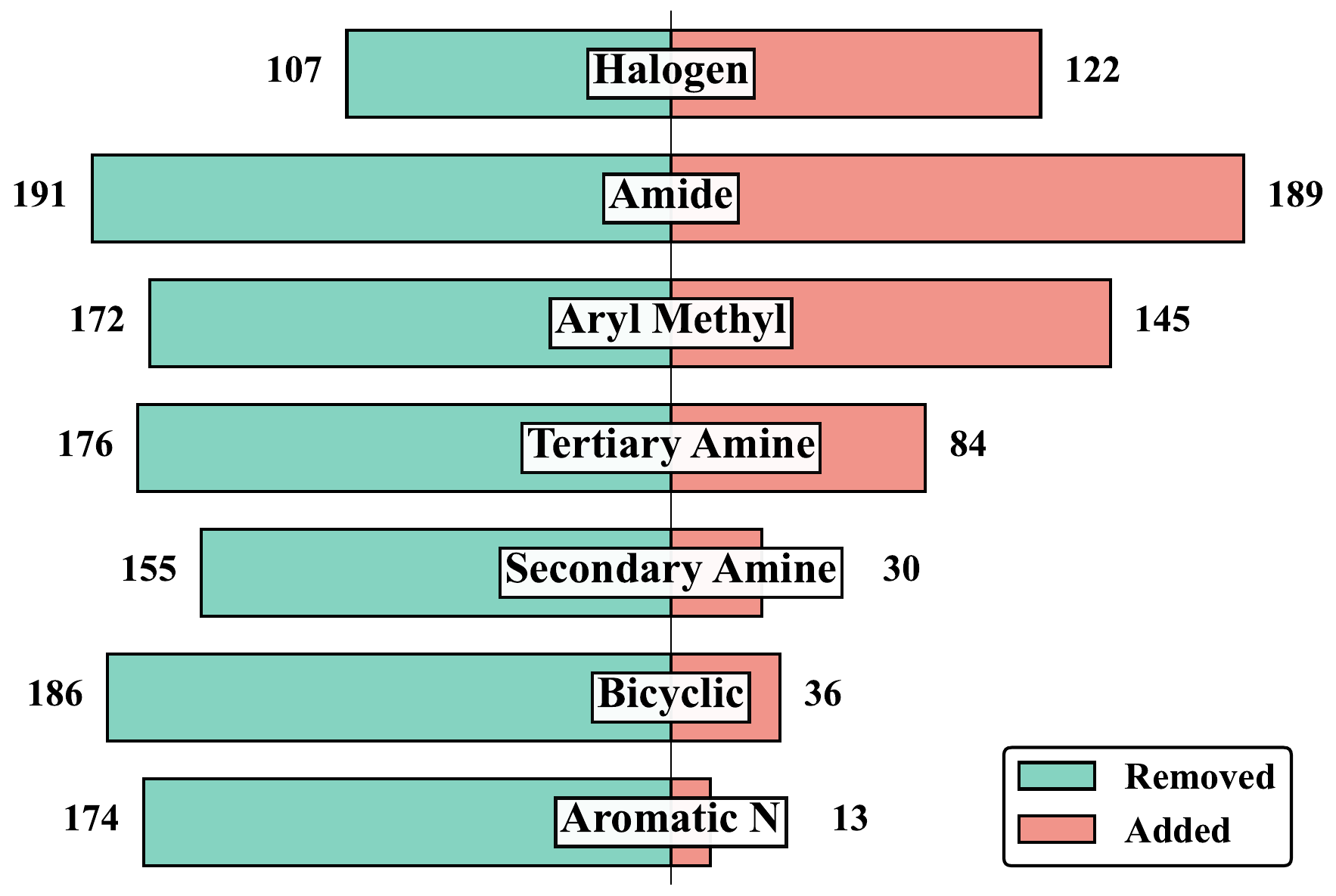}
    \caption{Functional Group Flux}
    \label{fig:fg}
\end{subfigure}
\hfill
\begin{subfigure}[b]{0.32\textwidth}
    \centering
    \includegraphics[height=3.3cm]{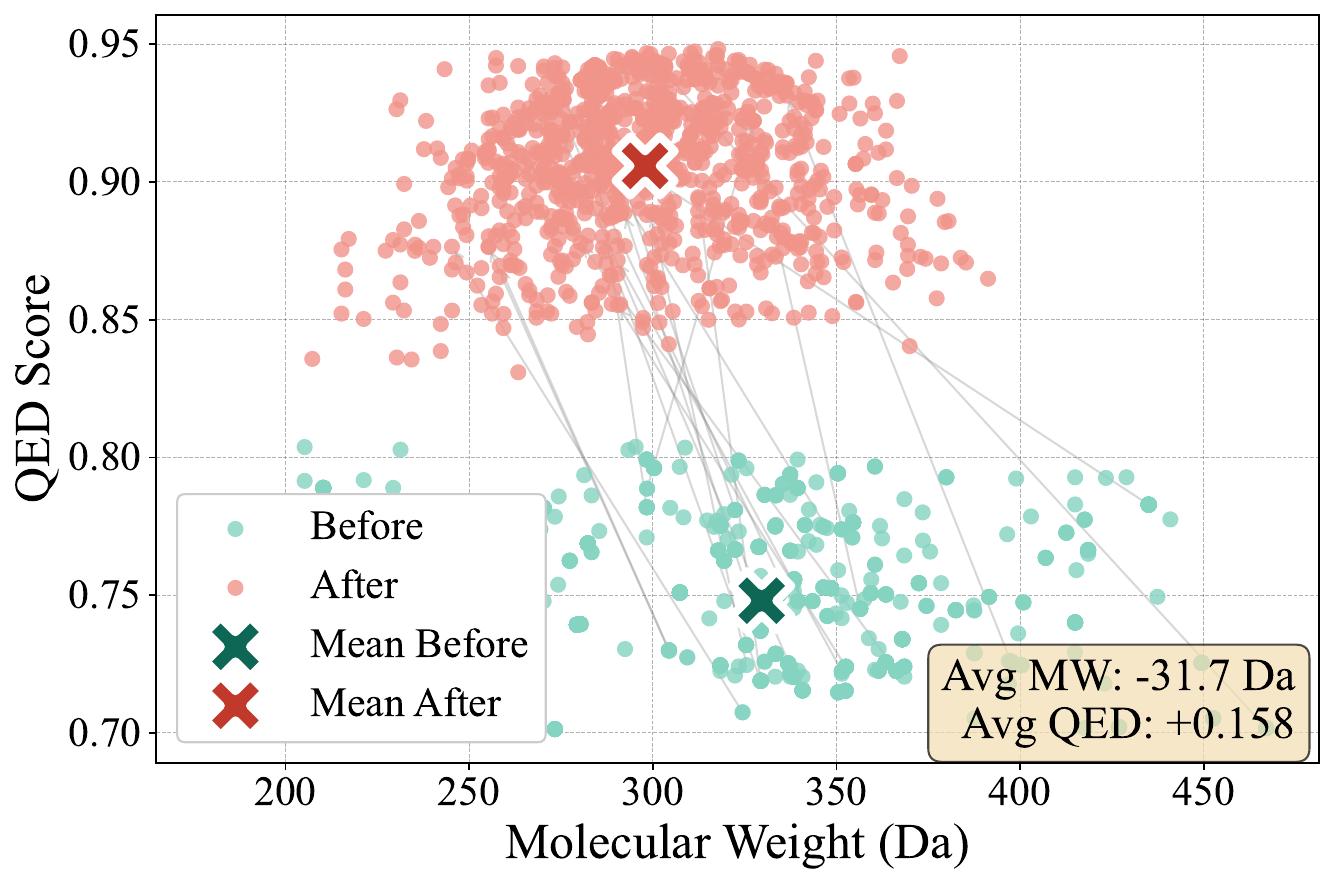}
    \caption{Chemical Space Trajectories}
    \label{fig:scatter}
\end{subfigure}
\vspace{-2mm}
\caption{\small\textbf{What the skill bank learns (QED task).}
(a) Distribution of scaffold modifications, where ring removal is the most frequent operation.
(b) Functional-group flux suggests a tendency to remove amine groups while adding halogen substitutions.
(c) Optimization trajectories indicate that QED improvements often coincide with reductions in molecular weight.}
\label{fig:analysis}
\vspace{-2mm}
\end{figure*}

\subsection{Analysis of Key Hyperparameters}
\label{sec:hyperparameters}

Figure~\ref{fig:sensitivity} analyzes the impact of critical hyperparameters on optimization performance. \textbf{(1) Importance of multi-turn optimization.} As shown in Figure~\ref{fig:traj_length}, allowing the agent to optimize over multiple turns yields substantial improvements over single-turn baselines. Performance gains saturate around $T{=}5$, indicating that this range provides an effective balance between optimization quality and computational cost. \textbf{(2) Data efficiency.} Figure~\ref{fig:train_size} shows that performance scales with training data size, but even 64 leads yield competitive results across tasks. This efficiency stems from the dual-memory system, which allows the agent to accumulate and reuse successful strategies across rollouts, amplifying the utility of limited training data. \textbf{(3) Robustness to similarity constraints.} Figure~\ref{fig:sim_threshold} shows that stricter similarity thresholds reduce success rates across all tasks, with harder targets (SA, JNK3) showing larger drops. Nevertheless, \method{} maintains reasonable performance, demonstrating the robustness of \method{}.

\subsection{What the Skill Bank Learns}
\label{sec:analysis}
To interpret the strategies captured by the evolving skill memory, we analyze learned skills on the QED task (Figure~\ref{fig:analysis}). The skill bank is dominated by scaffold-level transformations: ring removal accounts for 48.7\% of edits, followed by scaffold replacement (21.4\%) and ring addition (15.6\%) (Figure~\ref{fig:scaffold}). This tendency toward simplification is consistent with common drug-likeness guidelines that favor avoiding overly large or complex structures~\citep{lipinski2000drug,bickerton2012quantifying}. At the functional-group level, the learned skills more often remove amine-related motifs while slightly favoring halogen additions, with amide changes remaining roughly balanced (Figure~\ref{fig:fg}). Such edits reflect standard medicinal-chemistry levers used to tune drug-like properties during optimization~\citep{gleeson2008generation}. Consistent with these patterns, optimization trajectories shift toward lower molecular weight and higher QED (Avg MW $-31.7$ Da; Avg QED $+0.158$) (Figure~\ref{fig:scatter}). Overall, the skill bank captures reusable structure-level heuristics that move candidates toward more QED-favorable regions of chemical space.

\subsection{Ablation Study}
\label{sec:ablation}

We conduct an ablation study to quantify the contribution of each component, as summarized in Table~\ref{tab:ablation_study}. 
\textbf{(1) Multi-turn optimization is foundational.} Restricting the agent to single-turn optimization ($T{=}1$) substantially degrades success rate, from 91.0\% to 50.0\% on QED and from 96.0\% to 54.0\% on DRD2, highlighting the importance of iterative refinement with intermediate feedback. \textbf{(2) RL training is essential beyond a supervised prior.} Removing RL and using SFT-only reduces SR to 37.0\% (QED) and 20.0\% (DRD2), indicating that multi-turn optimization is necessary to realize consistent gains under a fixed oracle budget. 
\textbf{(3) SFT initialization is particularly critical for bioactivity.} Without SFT warm-start, DRD2 SR drops from 96.0\% to 34.5\%, suggesting that learning effective edits for challenging targets benefits from a strong chemical prior. 
\textbf{(4) Memory components are complementary.} Removing either Static Exemplar Memory or Evolving Skill Memory reduces SR on both tasks, and removing both causes the larger degradation, supporting that cold-start grounding and consolidated strategies provide additive benefits.

\input{tables/ablation}

%% file: tables/single_property.tex
\definecolor{novicecolor}{HTML}{D6FAE1}    
\definecolor{novicetag}{HTML}{2EAA85}    

\definecolor{apprenticecolor}{HTML}{D6F6FD}    
\definecolor{apprenticetag}{HTML}{3498DB}       

\definecolor{expertcolor}{HTML}{FDECDF}   
\definecolor{experttag}{HTML}{E86B5D}       
\definecolor{marlocolor}{HTML}{FCEAE7}

\definecolor{gold}{HTML}{FFD700}      
\definecolor{silver}{HTML}{C0C0C0}     
\definecolor{bronze}{HTML}{CD7F32}     

\newcommand{\first}[1]{\cellcolor{gold!30}\textbf{#1}}
\newcommand{\second}[1]{\cellcolor{silver!40}{#1}}
\newcommand{\third}[1]{\cellcolor{bronze!30}#1}

\begin{table*}[t!]
\centering
\setlength{\tabcolsep}{0pt}%
\renewcommand{\arraystretch}{0.9}
\caption{\small\textbf{Single-Property Optimization Results.} Methods are categorized by optimization paradigm: 
\textcolor{novicetag}{\textbf{Novice}}, \textcolor{apprenticetag}{\textbf{Apprentice}}, and \textcolor{experttag}{\textbf{Expert}}. 
SR (\%) denotes the success rate, Sim denotes the Tanimoto similarity to the original molecule (we require $\mathrm{Sim}\ge 0.4$), and RI denotes relative improvement. For SR, \colorbox{gold!30}{\textbf{gold}} = best, \colorbox{silver!40}{silver} = second, and \colorbox{bronze!30}{bronze} = third.}

\label{tab:single}
\vspace{-3pt}
\resizebox{\textwidth}{!}{
\begin{threeparttable}
\begin{tabular}{
   @{\hspace{4pt}}l@{\hspace{6pt}}
   l@{\hspace{10pt}}     
   @{\hspace{2pt}}c@{\hspace{4pt}}
   @{\hspace{2pt}}c@{\hspace{4pt}}
   @{\hspace{2pt}}c@{\hspace{2pt}}
   @{\hspace{5pt}}c@{\hspace{5pt}}      
   @{\hspace{0pt}}c@{\hspace{4pt}}
   @{\hspace{2pt}}c@{\hspace{4pt}}
   @{\hspace{2pt}}c@{\hspace{2pt}}
   @{\hspace{5pt}}c@{\hspace{5pt}}      
   @{\hspace{0pt}}c@{\hspace{4pt}}
   @{\hspace{2pt}}c@{\hspace{4pt}}
   @{\hspace{2pt}}c@{\hspace{2pt}}
   @{\hspace{5pt}}c@{\hspace{5pt}}      
   @{\hspace{0pt}}c@{\hspace{4pt}}
   @{\hspace{2pt}}c@{\hspace{4pt}}
   @{\hspace{2pt}}c@{\hspace{2pt}}
   @{\hspace{5pt}}c@{\hspace{5pt}}      
   @{\hspace{0pt}}c@{\hspace{4pt}}
   @{\hspace{2pt}}c@{\hspace{4pt}}
   @{\hspace{2pt}}c@{\hspace{0pt}}
}
\toprule
\multirow{2}{*}{\textbf{Method}} & \multirow{2}{*}{\textbf{Backbone}} & \multicolumn{3}{c}{\textbf{QED}} && \multicolumn{3}{c}{\textbf{plogP}} && \multicolumn{3}{c}{\textbf{SA}} && \multicolumn{3}{c}{\textbf{DRD2}} && \multicolumn{3}{c}{\textbf{JNK3}} \\
\cmidrule{3-5} \cmidrule{7-9} \cmidrule{11-13} \cmidrule{15-17} \cmidrule{19-21}
&& SR (\%) & Sim & RI && SR (\%) & Sim & RI && SR  (\%) & Sim & RI && SR (\%) & Sim & RI && SR (\%) & Sim & RI \\
\midrule

\multicolumn{21}{c}{\textcolor{novicetag}{\ding{182}} \textbf{\textcolor{novicetag}{Novice}}} \\
Graph-GA & N/A & 59.5 & 0.49 & 0.13 && \third{61.5} & 0.49 & 9.64 && \second{46.0} & 0.48 & 0.20 && 34.0 & 0.56 & 5.13 && 6.5 & 0.59 & 1.73 \\
QMO & GRU & 16.5 & 0.53 & 0.16 && 8.0 & 0.55 & 5.87 && 6.5 & 0.52 & 0.15 && 9.5 & 0.51 & 3.91 && 0.5 & 0.56 & 1.20 \\
Reinvent 4 & Transformer & 33.5 & 0.50 & 0.14 && 51.5 & 0.48 & 10.37 && 18.5 & 0.47 & 0.21 && 34.5 & 0.50 & 8.71 && 28.0 & 0.51 & 1.91 \\
\midrule

\multicolumn{21}{c}{\textcolor{apprenticetag}{\ding{183}} \textbf{\textcolor{apprenticetag}{Apprentice}}} \\
\addlinespace[3pt]
\rowcolor{apprenticecolor!40}
\multicolumn{21}{c}{\textit{Retrieval-based}} \\
Direct Retrieval & N/A & \third{68.5} & 0.43 & 0.21 && 53.5 & 0.43 & 13.13 && 38.0 & 0.44 & 0.30 && 28.5 & 0.43 & 8.10 && 21.0 & 0.44 & 2.45 \\
\addlinespace[3pt]
\rowcolor{apprenticecolor!40}
\multicolumn{21}{c}{\textit{Prompting \& SFT}} \\
\multirow{2}{*}{Direct Prompt} & Qwen2.5-1.5B & 53.0 & 0.66 & 0.18 && 38.0 & 0.65 & 12.48 && 13.0 & 0.69 & 0.16 && 13.5 & 0.64 & 4.92 && 19.5 & 0.66 & 2.23 \\
                               & Qwen2.5-3B   & 67.5 & 0.63 & 0.20 && 52.5 & 0.63 & 14.65 && 34.0 & 0.66 & 0.27 && 25.5 & 0.68 & 8.14 && 31.5 & 0.64 & 1.79 \\
\multirow{2}{*}{SFT-only}      & Qwen2.5-1.5B & 37.0 & 0.62 & 0.14 && 24.5 & 0.64 & 9.09  && 8.0  & 0.62 & 0.13 && 20.0 & 0.62 & 7.04 && 8.5 & 0.63 & 0.72 \\
                               & Qwen2.5-3B   & 48.0 & 0.61 & 0.17 && 23.5 & 0.63 & 9.67  && 10.5 & 0.62 & 0.16 && 23.0 & 0.64 & 6.70 && 10.5 & 0.64 & 0.81 \\
\addlinespace[3pt]                              
\rowcolor{apprenticecolor!40}
\multicolumn{21}{c}{\textit{Task-Specific LLM}} \\
MOLLEO & BioT5 & 15.0 & 0.68 & 0.08 && 12.5 & 0.70 & 3.37 && 8.0 & 0.81 & 0.02 && 4.0 & 0.78 & 0.63 && 10.0 & 0.88 & 0.04 \\
LlaSMol & Mistral-7B & 38.0 & 0.45 & 0.14 && 32.5 & 0.49 & 9.98 && 20.5 & 0.50 & 0.19 && 15.5 & 0.47 & 5.94 && 6.0 & 0.44 & 0.71 \\
ChemLLM & ChemLLM-7B & 69.0 & 0.65 & 0.21 && 61.5 & 0.64 & 19.86 && 35.5 & 0.69 & 0.27 && 46.0 & 0.64 & 11.71 && \second{44.0} & 0.65 & 2.48 \\
GeLLM$^3$O & Llama3.1-8B & 61.5 & 0.56 & 0.19 && 57.0 & 0.52 & 12.80 && 14.5 & 0.57 & 0.16 && \third{49.0} & 0.56 & 11.16 && 8.5 & 0.55 & 1.19 \\
PEIT-LLM & Llama3.1-8B & \second{82.5} & 0.47 & 0.23 && \second{81.5} & 0.49 & 15.41 && \third{45.0} & 0.50 & 0.31 && \second{50.5} & 0.50 & 11.86 && \third{41.0} & 0.50 & 2.51 \\
\midrule

\multicolumn{21}{c}{\textcolor{experttag}{\ding{184}} \textbf{\textcolor{experttag}{Expert}}} \\
\textbf{\method{} (Ours)} & Qwen2.5-1.5B & \first{91.0} & 0.47 & 0.24 && \first{100.0} & 0.47 & 19.44 && \first{63.5} & 0.45 & 0.34 && \first{96.0} & 0.49 & 16.96 && \first{98.5} & 0.47 & 8.87 \\
\bottomrule
\end{tabular}
\end{threeparttable}
}
\vspace{-4mm}
\end{table*}

%% file: tables/multi_properties.tex
\begin{table*}[t!]
\centering
\setlength{\tabcolsep}{0pt}%
\renewcommand{\arraystretch}{0.9}
\caption{\small\textbf{Multi-Property Optimization Results.} Methods are categorized by optimization paradigm: \textcolor{novicetag}{\textbf{Novice}}, \textcolor{apprenticetag}{\textbf{Apprentice}}, and \textcolor{experttag}{\textbf{Expert}}. SR (\%) denotes the success rate, Sim denotes the Tanimoto similarity to the original molecule (we require $\mathrm{Sim}\ge 0.4$), and RI denotes relative improvement. For SR (Success Rate), 
\colorbox{gold!30}{\textbf{gold}} = best, 
\colorbox{silver!40}{\underline{silver}} = second, 
\colorbox{bronze!30}{bronze} = third.}
\label{tab:multi}
\vspace{-3pt}
\resizebox{\textwidth}{!}{
\begin{threeparttable}
\begin{tabular}{
   @{\hspace{4pt}}l@{\hspace{6pt}}
   l@{\hspace{10pt}}     
   @{\hspace{2pt}}c@{\hspace{4pt}}
   @{\hspace{2pt}}c@{\hspace{4pt}}
   @{\hspace{2pt}}c@{\hspace{2pt}}
   @{\hspace{5pt}}c@{\hspace{5pt}}      
   @{\hspace{0pt}}c@{\hspace{4pt}}
   @{\hspace{2pt}}c@{\hspace{4pt}}
   @{\hspace{2pt}}c@{\hspace{2pt}}
   @{\hspace{5pt}}c@{\hspace{5pt}}      
   @{\hspace{0pt}}c@{\hspace{4pt}}
   @{\hspace{2pt}}c@{\hspace{4pt}}
   @{\hspace{2pt}}c@{\hspace{2pt}}
   @{\hspace{5pt}}c@{\hspace{5pt}}      
   @{\hspace{0pt}}c@{\hspace{4pt}}
   @{\hspace{2pt}}c@{\hspace{4pt}}
   @{\hspace{2pt}}c@{\hspace{2pt}}
   @{\hspace{5pt}}c@{\hspace{5pt}}      
   @{\hspace{0pt}}c@{\hspace{4pt}}
   @{\hspace{6pt}}c@{\hspace{0pt}}
   @{\hspace{0pt}}c@{\hspace{0pt}}
}
\toprule
\multirow{2}{*}{\textbf{Method}} & \multirow{2}{*}{\textbf{Backbone}} & \multicolumn{3}{c}{\textbf{QED+plogP}} && \multicolumn{3}{c}{\textbf{plogP+DRD2}} && \multicolumn{3}{c}{\textbf{QED+SA}} && \multicolumn{3}{c}{\textbf{DRD2+SA}} && \multicolumn{3}{c}{\textbf{ DRD2+QED+plogP}} \\
\cmidrule{3-5} \cmidrule{7-9} \cmidrule{11-13} \cmidrule{15-17} \cmidrule{19-21}
&& SR (\%) & Sim & RI && SR (\%) & Sim & RI && SR  (\%) & Sim & RI && SR (\%) & Sim & RI && SR (\%) & Sim & RI \\
\midrule

\multicolumn{21}{c}{\textcolor{novicetag}{\ding{182}} \textbf{\textcolor{novicetag}{Novice}}} \\
Graph-GA & N/A & 8.0 & 0.50 & 4.11 && 2.5 & 0.48 & 6.61 && 8.0 & 0.52 & 0.12 && 0.0 & 0.53 & 4.15 && 0.0 & 0.51 & 3.79 \\
QMO & GRU & 3.0 & 0.52 & 6.40 && 2.0 & 0.51 & 4.20 && 1.5 & 0.53 & 0.11 && 1.0 & 0.50 & 2.49 && 0.0 & 0.52 & 1.14 \\
Reinvent 4 & Transformer & 6.0 & 0.49 & 8.60 && \second{50.0} & 0.48 & 7.94 && 17.0 & 0.66 & 0.05 && \second{37.5} & 0.50 & 5.54 && 3.0 & 0.51 & 4.28 \\
\midrule

\multicolumn{21}{c}{\textcolor{apprenticetag}{\ding{183}} \textbf{\textcolor{apprenticetag}{Apprentice}}} \\
\rowcolor{apprenticecolor!40}
\addlinespace[3pt]
\multicolumn{21}{c}{\textit{Retrieval-based}} \\
Direct Retrieval & N/A & \second{24.0} & 0.43 & 4.51 && 21.5 & 0.44 & 7.52 && \third{31.0} & 0.43 & 0.25 && 23.5 & 0.43 & 2.64 && \second{8.0} & 0.43 & 3.72 \\
\addlinespace[3pt]
\rowcolor{apprenticecolor!40}
\multicolumn{21}{c}{\textit{Prompting \& SFT}} \\
\multirow{2}{*}{Direct Prompt} & Qwen2.5-1.5B & 10.0 & 0.58 & 3.49 && 8.0 & 0.55 & 8.98 && 24.0 & 0.49 & 0.22 && 6.0 & 0.59 & 3.59 && 2.0 & 0.54 & 4.60 \\
                               & Qwen2.5-3B   & 12.5 & 0.54 & 4.85 && 7.0 & 0.57 & 8.58 && 21.5 & 0.55 & 0.16 && 5.0 & 0.58 & 3.36 && 1.0 & 0.58 & 3.68 \\
\multirow{2}{*}{SFT-only}      & Qwen2.5-1.5B & 14.0 & 0.60 & 3.46 && 11.5 & 0.62 & 7.16 && 20.0 & 0.59 & 0.13 && 6.5 & 0.62 & 2.74 && 1.0 & 0.61 & 2.61 \\
                               & Qwen2.5-3B   & 18.0 & 0.59 & 2.89 && 7.5 & 0.62 & 7.86 && 21.0 & 0.60 & 0.14 && 7.0 & 0.61 & 2.41 && 1.5 & 0.61 &  2.83 \\
                               \addlinespace[3pt]
\rowcolor{apprenticecolor!40}
\multicolumn{21}{c}{\textit{Task-Specific LLM}} \\
LlaSMol & Mistral-7B & 15.0 & 0.49 & 3.58 && 8.0 & 0.46 & 9.16 && 21.0 & 0.48 & 0.19 && 16.0 & 0.47 & 3.79 && 1.5 & 0.47 & 3.59 \\
ChemLLM & ChemLLM-7B & 11.5 & 0.51 & 6.36 && 8.0 & 0.53 & 8.06 && 20.5 & 0.51 & 0.21 && 22.0 & 0.55 & 5.40 && 0.0 & 0.52 & 2.85 \\
GeLLM$^3$O & Llama3.1-8B & \third{19.5} & 0.60 & 2.89 && 16.0 & 0.58 & 7.41 && 22.5 & 0.57 & 0.14 && 9.0 & 0.59 & 2.76 && 0.0 & 0.61 & 2.92 \\
PEIT-LLM & Llama3.1-8B & 12.5 & 0.44 & 5.95 && \third{25.0} & 0.47 & 12.82 && \second{41.5} & 0.44 & 0.26 && \third{34.5} & 0.46 & 5.61 && \third{4.0} & 0.45 & 5.51 \\
\midrule
\multicolumn{21}{c}{\textcolor{experttag}{\ding{184}} \textbf{\textcolor{experttag}{Expert}}} \\
\textbf{\method{} (Ours)} & Qwen2.5-1.5B & \first{58.0} & 0.49 & 12.12 && \first{54.0} & 0.48 & 13.78 && \first{64.5} & 0.47 & 0.24 && \first{69.5} & 0.48 & 7.11 && \first{15.0} & 0.46 & 6.26 \\
\bottomrule
\end{tabular}
\end{threeparttable}
}
\vspace{-1mm}
\end{table*}

%% file: tables/ablation.tex
\definecolor{marlocolor}{HTML}{FCEAE7}  

\begin{table}[t]
\centering
\caption{\small \textbf{Ablation Study of Component Contributions.}}
\label{tab:ablation_study}
\vspace{-2mm}
\resizebox{\linewidth}{!}{%
\begin{tabular}{lcccccc}
\toprule
\multirow{2}{*}{\textbf{Model Configuration}} & \multicolumn{3}{c}{\textbf{QED}} & \multicolumn{3}{c}{\textbf{DRD2}} \\
\cmidrule(lr){2-4} \cmidrule(lr){5-7}
& SR (\%)  & Sim & RI & SR (\%)  & Sim & RI \\
\midrule
\rowcolor{marlocolor}
\textbf{\method{} (Full)} 
& \textbf{91.0} & \textbf{0.47} & \textbf{0.24}
& \textbf{96.0} & \textbf{0.49} & \textbf{16.97} \\
\midrule

\multicolumn{7}{c}{\textit{--- Ablation of Training Components ---}} \\
w/o SFT Initialization 
& 68.0 & 0.54 & 0.19
& 34.5 & 0.58 & 8.64 \\
w/o Multi-turn (T=1)
& 50.0  & 0.55 & 0.19
& 54.0 & 0.58  & 11.39 \\
w/o RL Training (SFT-only) 
& 37.0 & 0.62 & 0.14
& 20.0 & 0.62 & 7.04 \\
\midrule

\multicolumn{7}{c}{\textit{--- Ablation of Memory Components ---}} \\
w/o Static Exemplar Memory 
& 81.0 & 0.52 & 0.20
& 88.5 & 0.49 & 15.56 \\
w/o Evolving Skill Memory 
& 83.0 & 0.52 & 0.21
& 85.5 & 0.49 & 14.78 \\
w/o Both Memories
& 80.0 & 0.51 & 0.20
& 81.0 & 0.50 & 14.12 \\
\bottomrule
\end{tabular}%
}
\vspace{-4mm}
\end{table}

%% file: sec/6_conclusion.tex
\section{Conclusion}
\label{sec:conclusion}

We presented \method{}, a memory-augmented multi-turn agentic RL framework for sample-efficient molecular optimization. In our dual-memory system, Static Exemplar Memory provides cold-start grounding, while Evolving Skill Memory distills successful trajectories into reusable optimization strategies. Experiments demonstrate strong performance on both single-property (90\% SR) and multi-property (52\% SR) tasks using only 500 oracle calls, substantially outperforming existing methods. An analysis of the learned skill bank further suggests that \method{} captures chemically meaningful patterns (e.g., frequent ring removal and systematic amine reduction) that are consistent with established medicinal chemistry heuristics.  We believe this work takes a step toward more practical molecular optimization agents.

%% file: sec/7_limitations.tex
\section{Limitations}
First, our evaluation relies on computational oracles. These surrogates, while widely used in molecular optimization benchmarks, may not fully reflect outcomes measured by wet-lab assays or other higher-fidelity evaluations. Second, skill summarization currently depends on an external LLM (GPT-4o) to convert structured edit cards into natural language strategies. While effective, this introduces additional inference cost. Integrating summarization directly into the RL training loop in an end-to-end fashion is an important next step.

%% file: sec/X_suppl.tex
\section{LLM Usage}

We used Large Language Models (ChatGPT/Claude/Gemini) exclusively for grammatical correction in this manuscript. The LLMs played no role in research ideation, methodology, or scientific content generation. All technical contributions and scientific insights are original work by the authors.

\section{Baselines and Implementation Details}
\label{appendix:baselines}

\subsection{Baseline Methods}
We compare \method{} with baselines that cover two common optimization paradigms in Fig.~\ref{fig:evolution}: 
\textbf{(i) Novice} trial-and-error search and \textbf{(ii) Apprentice} methods that rely on external knowledge (retrieval, pretrained LLMs, or supervised fine-tuning).

\paragraph{Novice baselines.}
\begin{itemize}[leftmargin=*,topsep=0pt]
\item \textbf{Graph-GA}~\citep{jensen2019graph} is a graph-based genetic algorithm that iteratively mutates and recombines molecules with chemistry-aware operators, using oracle scores as selection signals.

\item \textbf{QMO}~\citep{hoffman2022optimizing} performs black-box optimization in the latent space of a pretrained molecular autoencoder, updating candidates via zeroth-order gradient estimates from oracle evaluations.

\item \textbf{Reinvent~4}~\citep{loeffler2024reinvent} is an on-policy RL method that trains a SMILES generator (e.g., Transformer) to maximize a scoring function under reward shaping.
\end{itemize}

\paragraph{Apprentice baselines.}
\begin{itemize}[leftmargin=*,topsep=0pt]
\item \textbf{Direct Retrieval} uses the same static retriever as \method{} to retrieve structurally similar molecules from the external database and selects the best candidate under the task objective, subject to the similarity constraint. This baseline isolates the effect of retrieval without learning.

\item \textbf{Direct Prompt} queries general-purpose instruction-tuned LLMs using the same task instruction as \method{} and treats each proposal independently.

\item \textbf{SFT-only} uses the same supervised data and training recipe as Stage~I of \method{}, but does not apply RL. At test time, it generates candidates by prompting the fine-tuned model without memory or policy optimization.

\item \textbf{Task-specific LLMs} include MOLLEO~\citep{wang2024efficient}, LlaSMol~\citep{yu2024llasmol}, ChemLLM~\citep{zhang2024chemllm}, PEIT-LLM~\citep{lin2024property}, and GeLLM$^3$O~\citep{dey2025mathtt}. These models are designed or fine-tuned for chemistry and/or molecular optimization settings. We follow the evaluation protocols described in their respective papers when applicable, while enforcing the same oracle-call budget and similarity constraint as in our setup.
\end{itemize}

\subsection{Implementation Details}
\paragraph{Budget and constraints.}
For each lead molecule, all methods are evaluated under the same oracle-call budget $B{=}500$ and the same similarity constraint (Tanimoto similarity $\ge 0.4$). An oracle call is counted whenever a candidate molecule is evaluated by the property oracle(s). We report the best feasible molecule found within the budget. For fair comparison, all methods are evaluated with the same constraints and budget, and we apply identical success criteria and metrics across methods (Appendix~\ref{appendix:metrics}).

\paragraph{Traditional methods.}
For Graph-GA, QMO, and Reinvent~4, we use official implementations when available and otherwise use widely adopted public re-implementations. We keep their default hyperparameters, and we modify the optimization loop to explicitly track oracle calls so that each method respects the same budget.

\paragraph{LLM-based methods.}
For Direct Prompt and task-specific LLM baselines, we use a shared prompting template (Appendix~\ref{appendix:sft}) and a common parsing and validation pipeline. Each model produces candidate SMILES strings, which are canonicalized and checked for validity. Valid candidates are then evaluated by the oracle and counted toward the budget. We repeat the generation until the budget is exhausted. We use the temperature $\tau{=}0.9$ for sampling.

\section{Evaluation Metrics Details}
\label{appendix:metrics}

\subsection{Success Rate (SR)}
Success Rate (SR) is the percentage of lead molecules for which the method finds at least one \emph{feasible} optimized molecule $m'$ within the oracle budget. A molecule is feasible if it satisfies the similarity constraint $\mathrm{sim}(m,m') \ge 0.4$ and the task-specific success criterion in Table~\ref{tab:success_criteria}. For multi-property tasks, a run is counted as successful only if \emph{all} target properties meet their respective thresholds.

\begin{table}[t]
\centering
\caption{Task-specific success criteria. For multi-property tasks, $\Delta F = F(m')-F(m)$ is measured relative to the lead molecule $m$.}
\resizebox{\linewidth}{!}{
\begin{tabular}{lcc}
\toprule
\textbf{Property} & \textbf{Single-property success} & \textbf{Multi-property success} \\
\midrule
QED   & $F_{\text{QED}}(m') \ge 0.9$ & $\Delta F_{\text{QED}} \ge 0.1$ \\
plogP & $F_{\text{plogP}}(m') \ge 2.0$ & $\Delta F_{\text{plogP}} \ge 1.0$ \\
JNK3  & $F_{\text{JNK3}}(m') \ge 0.1$ & $\Delta F_{\text{JNK3}} \ge 0.1$ \\
DRD2  & $F_{\text{DRD2}}(m') \ge 0.8$ & $\Delta F_{\text{DRD2}} \ge 0.5$ \\
SA    & $F_{\text{SA}}(m') \le -2.5$   & $\Delta F_{\text{SA}} \le -0.5$ \\
\bottomrule
\end{tabular}}
\label{tab:success_criteria}
\end{table}

\subsection{Similarity (Sim)}
We report the average Tanimoto similarity between each lead molecule $m_i$ and the best molecule returned by the method $m'_i$:
\begin{align}
\text{Sim} = \frac{1}{N}\sum_{i=1}^{N} \mathrm{sim}(m_i, m'_i),
\end{align}
where $N{=}200$. If a method fails to return any valid candidate that satisfies the constraints for lead $m_i$, we set $m'_i = m_i$ for that instance. This convention avoids dropping failures when aggregating results. As a consequence, Sim should be interpreted jointly with SR, since a method that frequently fails may obtain a high Sim by defaulting to the lead molecule.

\subsection{Relative Improvement (RI)}
Relative Improvement (RI) measures the average normalized improvement over the target properties. For a task with $n$ properties, we compute:
\begin{align}
\text{RI} = \frac{1}{n}\sum_{j=1}^{n} \mathrm{sgn}(w_j)\cdot \frac{F_j(m')-F_j(m)}{|F_j(m)|},
\end{align}
where $\mathrm{sgn}(w_j){=}+1$ for properties to maximize (QED, plogP, JNK3, DRD2) and $\mathrm{sgn}(w_j){=}-1$ for properties to minimize (SA). The absolute value in the denominator handles properties that may take negative values (e.g., plogP). If optimization fails (i.e., no feasible molecule is found within the budget), we set RI to 0 for that instance.

\section{Details of Supervised Fine-Tuning}
\label{appendix:sft}

This section describes the supervised fine-tuning (SFT) stage used to initialize the policy in \method{}.

\subsection{Training Data Construction}
We follow the task setup of GeLLM$^3$O~\citep{dey2025mathtt} and build SFT pairs from the molecule-pair dataset of Chen et al.~\citep{chen2021deep}. Each example is a pair $(M_x, M_y)$ that corresponds to a local, single-fragment modification. We filter pairs to encourage similarity-preserving edits by retaining only those with Tanimoto similarity $\ge 0.6$. For each retained pair, we compute task-relevant properties (QED, plogP, JNK3, DRD2, SA) and keep examples that show a meaningful improvement for the specified objective. Each pair is then converted into an instruction--response example by using $M_x$ as the input molecule and $M_y$ as the target output.

\subsection{Fine-Tuning Setup}
We fine-tune \textbf{Qwen2.5-1.5B-Instruct} using LoRA~\citep{hu2022lora} for parameter-efficient adaptation. Unless otherwise stated, we use LoRA rank $r{=}16$ and $\alpha{=}32$, and train for \textbf{10 epochs}. The resulting model is used as the SFT initialization for the policy $\pi_\theta$ in \method{}.

\subsection{Prompt Template}
We use the unified prompt template below for SFT data formatting and for LLM-based baselines to ensure consistent task specification.
\begin{tcolorbox}[
  colback=myblue!10!white,
  colframe=myblue!75!black,
  title=Unified Prompt Template,
  boxrule=0.6pt,
  arc=2mm,
  left=1mm,right=1mm,top=1mm,bottom=1mm,
  breakable
]
\begin{Verbatim}[
  fontsize=\scriptsize,
  breaklines,
  breakanywhere,
  breaksymbolleft={},
  breaksymbolright={}
]
You are an expert medicinal chemist specializing in molecular optimization. You understand how structural modifications affect key molecular properties including drug-likeness, lipophilicity, synthetic accessibility, and target inhibition activities.

Your task is to modify the given molecule to adjust the specified molecular properties while keeping structural changes as minimal as possible. The modified molecule should maintain a structural similarity of at least 0.6 with the original molecule.

Input molecule: <SMILES> {input_smiles} </SMILES>
Requested modifications: {property_description}

Please provide the optimized molecule in SMILES format, wrapped in <SMILES> </SMILES> tags.
\end{Verbatim}
\end{tcolorbox}

The \texttt{\{property\_description\}} field is instantiated according to the task objective:
\begin{itemize}[leftmargin=*,noitemsep,topsep=0pt]
\item QED: increase drug-likeness (QED)
\item LogP: increase lipophilicity (LogP)
\item JNK3/DRD2: increase inhibition probability
\item SA: decrease synthetic accessibility score (lower is better)
\item Multi-property: combine the above objectives with conjunctions
\end{itemize}

\subsection{Why high-similarity pairs for SFT.}
In Stage~I, our goal is to teach the LLM the SMILES syntax and to perform controlled, local edits rather than large scaffold jumps. We therefore construct SFT examples from molecule pairs with high structural overlap (Tanimoto similarity $\ge 0.6$). This design encourages the model to learn ``small but valid'' modifications that preserve the core structure. As a consequence, the \textbf{SFT-only} baseline tends to produce outputs with consistently high similarity in evaluation: the model is trained to stay close to the input molecule, which naturally raises Sim even when property improvements are limited.

\section{Environment and Reward Details}
\label{appendix:env_reward}

\subsection{Environment Interface and Termination}
\label{appendix:env_interface}

We formulate molecular optimization as a multi-turn interaction between an LLM agent and a molecule-editing environment. 
At turn $t$, the agent proposes a candidate molecule $m_{t+1}$ in SMILES form; the environment parses and validates it, evaluates oracle properties, and returns a scalar reward $r_t$ together with textual feedback for the next turn.

A rollout terminates when any of the following conditions is met: 
(1) the maximum trajectory length $T$ is reached; 
(2) a molecule satisfying the task-specific success criterion is found (under the similarity constraint).

\subsection{Reward Computation}
\label{appendix:reward}

The reward is designed to (i) discourage invalid or trivial edits, (ii) enforce the similarity constraint, and (iii) provide dense, step-wise learning signals from oracle feedback. 
Algorithm~\ref{alg:reward} summarizes the computation.

\begin{algorithm}[t]
\caption{Step-wise reward computation.}
\label{alg:reward}
\begin{algorithmic}[1]
\Require current molecule $m_t$; proposed molecule (SMILES) $\tilde{m}_{t+1}$; similarity threshold $\gamma$; target property oracle $F$ with direction $\text{sgn}(w_F)\in\{+1,-1\}$
\Ensure reward $r_t$
\State Parse $\tilde{m}_{t+1}$ into a molecule $m_{t+1}$; if parsing fails, return $r_t \leftarrow -0.5$
\State Canonicalize $m_t$ and $m_{t+1}$; if $m_{t+1}=m_t$ (no-op), return $r_t \leftarrow -0.3$
\State Compute similarity $\mathrm{sim}\leftarrow \mathrm{sim}(m_0,m_{t+1})$; if $\mathrm{sim}<\gamma$, return $r_t \leftarrow -2(\gamma-\mathrm{sim})$
\State Compute property change $\Delta F \leftarrow F(m_{t+1})-F(m_t)$
\If{$\text{sgn}(w_F)\cdot \Delta F > 0$} 
    \State $r_t \leftarrow 5|\Delta F|$ \Comment{property improves}
\Else
    \State $r_t \leftarrow -|\Delta F|$ \Comment{property degrades}
\EndIf
\end{algorithmic}
\end{algorithm}

We use asymmetric scaling so that genuine improvements receive a stronger learning signal than small degradations, which helps avoid overly conservative behavior under a strict similarity constraint.

\subsection{Implementation Details.} 
Tanimoto similarity is computed using Morgan fingerprints (radius $=2$, 2048 bits). 
When the task involves minimizing a property (e.g., SA), we set $\text{sgn}(w_F)=-1$ so that decreases count as improvements. 
All oracle evaluations are performed only after passing validity and similarity checks. We count one oracle call whenever a \emph{valid} candidate molecule passes parsing/canonicalization and is evaluated by the property oracle(s). Invalid SMILES and similarity-violating candidates are rejected before oracle evaluation and therefore do not consume the oracle-call budget.

\section{Hyperparameter Settings}
\label{appendix:hyperparameters}

\subsection{Training Configuration}
\label{appendix:train_hparams}
We train \method{} in two stages. 
We first apply supervised fine-tuning (SFT) to Qwen2.5-1.5B-Instruct with LoRA ($r{=}16$, $\alpha{=}32$) for 10 epochs to teach valid, similarity-preserving SMILES edits.
Starting from the SFT checkpoint, we further optimize the multi-turn policy using PPO for 100 update steps with learning rate $5{\times}10^{-5}$, minibatch size 32, and standard clipping/GAE settings ($\epsilon{=}0.2$, $\gamma_{\text{rl}}{=}0.99$, $\lambda{=}0.95$).
Each PPO iteration collects rollouts with horizon $T{=}5$ from 128 training leads, with 16 rollouts per lead and a similarity constraint $\gamma{=}0.4$. All experiments are run on 2$\times$H100 GPUs with maximum sequence length 4096.

\begin{table*}[t]
\centering
\caption{\small Training hyperparameters for \method{}.}
\label{tab:train_hparams}
\small
\begin{tabular}{llc}
\toprule
\textbf{Category} & \textbf{Parameter} & \textbf{Value} \\
\midrule
\multirow{5}{*}{Backbone \& SFT}
 & Base model & Qwen2.5-1.5B-Instruct \\
 & SFT epochs & 10 \\
 & LoRA rank $r$ & 16 \\
 & LoRA $\alpha$ & 32 \\
\midrule
\multirow{8}{*}{PPO}
 & Training steps & 100 \\
 & PPO learning rate & $5 \times 10^{-5}$ \\
 & PPO minibatch size & 32 \\
 & Clip ratio $\epsilon$ & 0.2 \\
 & Discount factor $\gamma_{\text{rl}}$ & 0.99 \\
 & GAE $\lambda$ & 0.95 \\
 & Micro batch size / GPU & 2 \\
 & Max sequence length & 4096 \\
\midrule
\multirow{5}{*}{Rollouts \& Env}
 & Max turns per rollout $T$ & 5 \\
 & Training leads per iteration & 128 \\
 & Rollouts per lead & 16 \\
 & Similarity threshold $\gamma$ & 0.4 \\
 & GPUs & 2$\times$H100 \\
\bottomrule
\end{tabular}
\end{table*}

\subsection{Evaluation Configuration}
\label{appendix:eval_hparams}

At test time, each lead molecule is optimized under a fixed oracle-call budget of $B{=}500$ and the same similarity threshold $\gamma{=}0.4$.
We run a search procedure for $G{=}20$ iterations.
In each iteration, the agent samples $N{=}32$ rollouts with horizon $T{=}5$ to propose candidate molecules, and we keep the best feasible molecule seen so far as the current incumbent.
To encourage diversity when improvements slow down, we use a temperature schedule across iterations:
$\tau_g = \min(\tau_0 + g\Delta\tau,\ \tau_{\max})$ with $\tau_0{=}0.9$, $\Delta\tau{=}0.1$, and $\tau_{\max}{=}2.0$.

\begin{table*}[t]
\centering
\caption{\small Inference hyperparameters for \method{}.}
\label{tab:inference_hyperparameters}
\small
\begin{tabular}{llc}
\toprule
\textbf{Category} & \textbf{Parameter} & \textbf{Value} \\
\midrule
\multirow{3}{*}{Budget \& constraints}
 & Oracle-call budget per lead $B$ & 500 \\
 & Similarity threshold $\gamma$ & 0.4 \\
 & Search iterations (generations) $G$ & 20 \\
\midrule
\multirow{2}{*}{Candidate generation}
 & Rollouts per iteration $N$ & 32 \\
 & Max turns per rollout $T$ & 5 \\
\midrule
\multirow{3}{*}{Decoding diversity}
 & Base temperature $\tau_{0}$ & 0.9 \\
 & Temperature increment $\Delta\tau$ & 0.1 \\
 & Max temperature $\tau_{\max}$ & 2.0 \\
\bottomrule
\end{tabular}
\vspace{-2mm}
\end{table*}

\section{Static Exemplar Memory Implementation}
\label{appendix:retrieval}

\paragraph{Database Construction.}
We build the static exemplar bank from ChEMBL~\citep{zdrazil2024chembl} (2.8M molecules). 
For each molecule, we precompute and store (i) oracle-relevant properties (QED, LogP, SA, and target activity scores such as JNK3/DRD2), 
(ii) an ECFP4 fingerprint (radius $=2$, 2048-bit) normalized for FAISS L2 search, and 
(iii) a binary fingerprint for fast Tanimoto computation.
We store the metadata in SQLite and build a FAISS IVF index (nlist$=1689$), with an index size of $\sim$22GB.

\paragraph{Retrieval Pipeline.}
At turn $t$, we retrieve exemplars based on the current molecule $m_t$, while enforcing the similarity constraint with respect to the original lead $m_0$:
\begin{enumerate}[leftmargin=*,nosep]
    \item \textbf{ANN recall.} Query FAISS with the normalized ECFP4 fingerprint of $m_t$ to obtain a candidate pool.
    \item \textbf{Lead-based filtering.} Compute Tanimoto similarity between each candidate and the lead $m_0$ (using binary fingerprints) and keep only those satisfying the lead similarity constraint.
    \item \textbf{Objective-aware ranking.} Rank the remaining candidates by the target objective and return the top-$K$ molecules as exemplars.
\end{enumerate}
This setup uses $m_t$ to stay in a relevant neighborhood, and uses $m_0$ to respect the lead-preserving constraint.

\paragraph{When retrieval is triggered.}
We do not retrieve at every turn. Instead, retrieval is triggered only when the optimization stalls:
if the agent fails to improve the target objective for \emph{two consecutive turns}, we query the exemplar bank and provide a small set of high-scoring, lead-similar references.

\paragraph{How exemplars are presented to the agent.}
Retrieved exemplars are appended to the observation as a compact reference block:

\begin{tcolorbox}[
  colback=myblue!8!white,
  colframe=myblue!75!black,
  title=Retrieved Template,
  boxrule=0.6pt,
  arc=2mm,
  left=1mm,right=1mm,top=1mm,bottom=1mm,
  breakable
]
\begin{Verbatim}[
  fontsize=\scriptsize,
  breaklines,
  breakanywhere,
  breaksymbolleft={},
  breaksymbolright={}
]
=== SIMILAR HIGH-SCORING MOLECULES FOR REFERENCE ===
Here are K similar molecules with high target scores (higher is better):

1. SMILES: CC1=CC=C(C=C1)NC(=O)C2=CC=CC=C2 
    target score: 0.892 
    Similarity to original lead: 0.654 

2. SMILES: 
    ... 
    ... 

Learn from structural patterns, but do not copy directly.
\end{Verbatim}
\end{tcolorbox}

\paragraph{Efficiency Notes.}
In our optimized implementation, retrieving the nearest-neighbor candidate set for a given query molecule takes about \textbf{4 ms} (measured with the FAISS index resident on GPU).
To keep retrieval fast, we (i) load the FAISS index on GPU, (ii) cache fingerprints to avoid recomputation,
(iii) compute Tanimoto similarity in batch with precomputed binary fingerprints, and (iv) tune IVF search parameters (e.g., \texttt{nprobe}) to balance latency and recall.

\section{Evolving Skill Memory Implementation}
\label{appendix:skill}

\paragraph{Skill Representation.}
We store each learned strategy as a \emph{skill card}. 
A card contains one short, reusable instruction in natural language, together with lightweight evidence so it can be retrieved and trusted later.
Concretely, each card records:
\begin{itemize}[leftmargin=*,nosep]
    \item \textbf{Skill text}: one actionable sentence (Sec.~\ref{sec:memory}), written in the form
    \emph{[Action] [What] [Where (if clear)] to [Effect]}.
    \item \textbf{Source edit}: the SMILES pair $(m_t, m_{t+1})$ and the observed improvement signal.
    \item \textbf{Edit summary}: an MCS-based decomposition of what changed (added/removed/replaced fragments),
    plus scaffold and functional-group changes detected by RDKit.
    \item \textbf{Retrieval keys}: an ECFP4 fingerprint and a functional-group tag set for the \emph{source} molecule.
\end{itemize}

\paragraph{Skill acquisition from rollouts.}
During training, we harvest skills from multi-turn rollouts by keeping only \emph{meaningful improving edits}.
For each step, we compare the oracle score before and after the edit and keep transitions that yield a clear improvement.
For bookkeeping, we also merge duplicates (e.g., identical SMILES pairs or near-identical edit patterns) and retain the best instance.

\paragraph{Turning edits into a reusable sentence.}
The structured edit information above is useful for retrieval, but it is not convenient for an LLM agent to apply directly.
We therefore use an external summarizer (GPT-4o) to rewrite each retained edit into \emph{one} strategy sentence.
The summarizer is given the before/after molecules and the detected fragment / functional-group changes, and is asked to produce a concise, actionable rule. We use the following prompt to convert each structured edit card into one reusable strategy sentence.

\begin{tcolorbox}[
  colback=myblue!10!white,
  colframe=myblue!75!black,
  title=Prompt Template of Summarizer Agent,
  boxrule=0.6pt,
  arc=2mm,
  left=1mm,right=1mm,top=1mm,bottom=1mm,
  breakable
]
\begin{Verbatim}[
  fontsize=\scriptsize,
  breaklines,
  breakanywhere,
  breaksymbolleft={},
  breaksymbolright={}
]
SUMMARIZER_PROMPT = """Analyze this molecular transformation for {task} optimization:

=== Molecules ===
Before: {before_smiles}
After:  {after_smiles}
Score:  {score_before:.3f} -> {score_after:.3f} ({score_delta:+.3f})

=== MCS Analysis ===
Modification: {modification_type}
- Removed: {removed_fragment}
- Added:   {added_fragment}

=== Scaffold Analysis ===
Before Scaffold: {before_scaffold}
After Scaffold:  {after_scaffold}
Scaffold Type: {scaffold_type}

=== Functional Group Changes ===
- Removed: {fg_removed}
- Added:   {fg_added}

=== Property Changes ===
- MW: {mw_change:+.1f} Da | Rings: {ring_changes:+d}
- PSA: {psa_change:+.1f} A2 | HBD: {hbd_change:+d} | HBA: {hba_change:+d}

Result: {result}

=== Task ===
Generate ONE actionable strategy sentence following this format:

CONSTRAINTS:
1. Focus on the 1-2 MOST IMPORTANT functional group changes
2. Format: "[Action] [What] [Where (if clear)] to [Effect]"
3. If scaffold_type is "scaffold_hop", mention the core change (e.g., "Replace benzene with pyridine").
4. Use the Removed/Added Fragment from MCS for precise description.
5. If location is clear from MCS, specify it (e.g., "on the aromatic ring").

EXAMPLES:
- "Replace benzene core with pyridine to improve water solubility and {task}." (scaffold_hop)
- "Add fluorine (-F) to the aromatic ring to enhance metabolic stability." (addition)
- "Remove the sulfonamide group from aromatic ring to reduce polar surface area." (removal)
- "Replace methoxy (-OCH3) with fluorine (-F) to decrease MW and improve {task}." (replacement)

Focus on: WHAT changed, WHERE (if clear from MCS), and WHY it improves {task}:"""
\end{Verbatim}
\end{tcolorbox}

Example outputs look like:
\begin{itemize}[leftmargin=*,nosep]
    \item ``Replace an aromatic methoxy with fluorine to reduce MW and improve QED.''
    \item ``Remove a tertiary amine side chain to reduce polarity while keeping the core scaffold.''
\end{itemize}

\paragraph{Skill Retrieval.}
At test time, we retrieve skills that are relevant to the current molecule in two complementary ways:
(1) \emph{structure-based} retrieval using fingerprint similarity, and 
(2) \emph{feature-based} retrieval using overlap in functional-group tags.
We then take a small set of top matches and present them as high-level hints.

\paragraph{When skills are used.}
Similar to exemplar retrieval, we do not inject skills at every turn.
If the agent makes no progress for \emph{two consecutive turns}, we retrieve a few relevant skills and add them to the next observation.

\paragraph{How retrieved skills are presented.}
Retrieved skills are appended as a compact hint block:

\begin{tcolorbox}[
  colback=myblue!8!white,
  colframe=myblue!75!black,
  title=Retrieved Skills,
  boxrule=0.6pt,
  arc=2mm,
  left=1mm,right=1mm,top=1mm,bottom=1mm,
  breakable
]
\begin{Verbatim}[
  fontsize=\scriptsize,
  breaklines,
  breakanywhere,
  breaksymbolleft={},
  breaksymbolright={}
]
=== Potential Useful Strategies for qed ===
1. Replace benzene core with pyridine to improve water solubility and qed.
2. Add fluorine (-F) to the aromatic ring to enhance metabolic stability.
3. Remove the sulfonamide group from aromatic ring to reduce polar surface area.
\end{Verbatim}
\end{tcolorbox}

\paragraph{Capacity Control.}
We cap the skill bank at 1{,}000 entries to keep it actively refreshed. 
When new skills are added beyond this limit, we apply a simple survival-of-the-fittest rule: we rank candidate skill cards by their improvement magnitude (score delta $\Delta$) and retain the top 1{,}000, discarding the rest. 
This favors high-impact skills while continuously removing weaker or redundant ones.

\paragraph{Summarizer Cost.}
Skill summarizer uses GPT-4o and incurs additional inference overhead. 
In our runs, summarizing skills for a single optimization task costs on the order of \$10 in API usage.

\section{Computational Cost Analysis}
\label{appendix:cost}

Table~\ref{tab:computational_cost} summarizes the required GPU time of representative baselines and \method{} on 2$\times$H100. 
A main difference is what can be reused across different molecules. Offline instruction tuning is typically a one-time cost, while online RL methods (like Reinvent~4) must be re-trained for each new lead molecule.

For \method{}, training has two parts. We first run a one-time SFT stage (10 hours) that teaches similarity-preserving SMILES edits and is reused across all tasks. We then run task-specific policy optimization (4 hours) for each objective. Evaluation on 200 test leads with an oracle-call budget of 500 per lead takes about 3 hours per task with batched rollouts.

\begin{table}[h]
\centering
\caption{
    \small \textbf{Computational cost comparison.}
    GPU hours are reported on 2$\times$H100.
    *Online RL baseline.
}
\label{tab:computational_cost}
\vspace{-6pt}
\resizebox{\linewidth}{!}{
\begin{tabular}{lcccc}
\toprule
\textbf{Method} & \textbf{Model} & \textbf{One-time} & \textbf{Per-task} & \textbf{Per-task}  \\
& \textbf{Size} & \textbf{Training} & \textbf{Training} & \textbf{Inference}  \\
\midrule
Reinvent~4* & --  & -- & 14h (online) & --  \\
GeLLM$^3$O  & 8B  & 48h (SFT) & -- & 1h  \\
\midrule
\textbf{\method{}} & 1.5B & 10h (SFT) & 4h & 3h  \\
\bottomrule
\end{tabular}}
\vspace{-6pt}
\end{table}

Overall, \method{} keeps per-task cost modest: after a reusable SFT checkpoint, task adaptation requires a short policy-optimization phase and batched multi-turn inference. Using a 1.5B backbone also lowers the hardware barrier compared to 7--8B LLM baselines.

\section{Case Studies}
\label{sec:case_study_skills}

We present two types of case studies. First, Fig.~\ref{fig:skill_cases} shows representative \emph{skill cards} distilled from successful transitions. These examples highlight that the skill bank captures insightful, actionable edit patterns rather than memorizing entire molecules. Second, we include a full multi-turn rollout example to illustrate how \method{} uses retrieved skills and exemplars.

\begin{figure*}[b]
\vspace{-2mm}
\centering
\begin{minipage}[t]{0.49\textwidth}
\skillcard
{DRD2 Skill 1}
{Replace the imidazole tail with a propyl--fluorophenyl group to reduce polarity and improve DRD2 affinity.}
{N\#Cc1c(F)cccc1N1CCN(CC2=CN=CNC2)CC1}
{N\#Cc1c(F)cccc1N1CCN(CCCc2ccc(F)cc2)CC1}
{0.01 $\rightarrow$ 0.99}
{\textbf{Hydrophobic tailoring:} A polar heterocycle is replaced by a more hydrophobic fluorophenyl tail, which is consistent with improved fit to hydrophobic regions in DRD2 ligands.}
\end{minipage}\hfill
\begin{minipage}[t]{0.49\textwidth}
\skillcard
{DRD2 Skill}
{Replace the small dimethylamine group with a fluorophenethyl--piperazine tail to introduce a DRD2-relevant pharmacophore.}
{CN(C)Cc1ccc(N=Nc2ccc3c(c2)CCCN3C)cc1}
{Cc1ccc(N=Nc2ccc3c(c2)CCCN3C)cc1N1CCN(CCc2ccc(F)cc2)CC1}
{0.03 $\rightarrow$ 1.00}
{\textbf{Pharmacophore grafting:} Adding a fluorophenethyl--piperazine tail introduces a stronger hydrophobic/basic motif, which is consistent with the score jump observed for DRD2.}
\end{minipage}

\vspace{2mm}

\begin{minipage}[t]{0.49\textwidth}
\skillcard
{QED Skill 1}
{Replace the furan ring with a cyclopropane group to reduce polar surface area and improve QED.}
{Cc1ccoc1C(=O)NCc1nc(-c2ccccc2)n[nH]1}
{CC1(C)CC1C(=O)NCc1nc(-c2ccccc2)n[nH]1}
{0.769 $\rightarrow$ 0.894}
{\textbf{Bioisosteric swap:} Replacing a heteroaromatic furan with an sp$^3$ cyclopropane reduces PSA ($-13.14$) while keeping the rest of the scaffold unchanged.}
\end{minipage}\hfill
\begin{minipage}[t]{0.49\textwidth}
\skillcard
{QED Skill 2}
{Replace the fluorobenzene (-C6H4F) group with an isopropyl group (-CH(CH3)2) to reduce molecular weight.}
{CCn1c(=O)n(CC(=O)NCc2ccc(F)c(C)c2)c2ccccc21}
{CCn1c(=O)n(CC(=O)NCC(C)C)c2ccccc21}
{0.775 $\rightarrow$ 0.901}
{\textbf{Side-chain simplification:} Removing an aromatic side chain reduces molecular weight ($-66$ Da), a change that often correlates with higher QED and is reflected in the score increase here.}
\end{minipage}

\vspace{-2mm}
\caption{\small \textbf{Case studies of learned skills.}
(Top) DRD2 examples highlighting tail substitution and motif augmentation.
(Bottom) QED examples highlighting bioisosteric replacement and side-chain simplification.
Each card shows the distilled strategy, the before/after edit, and the corresponding score change.}
\label{fig:skill_cases}
\vspace{-4mm}
\end{figure*}

\begin{figure*}[t]
\vspace{-2mm}
\centering
\includegraphics[width=\linewidth]{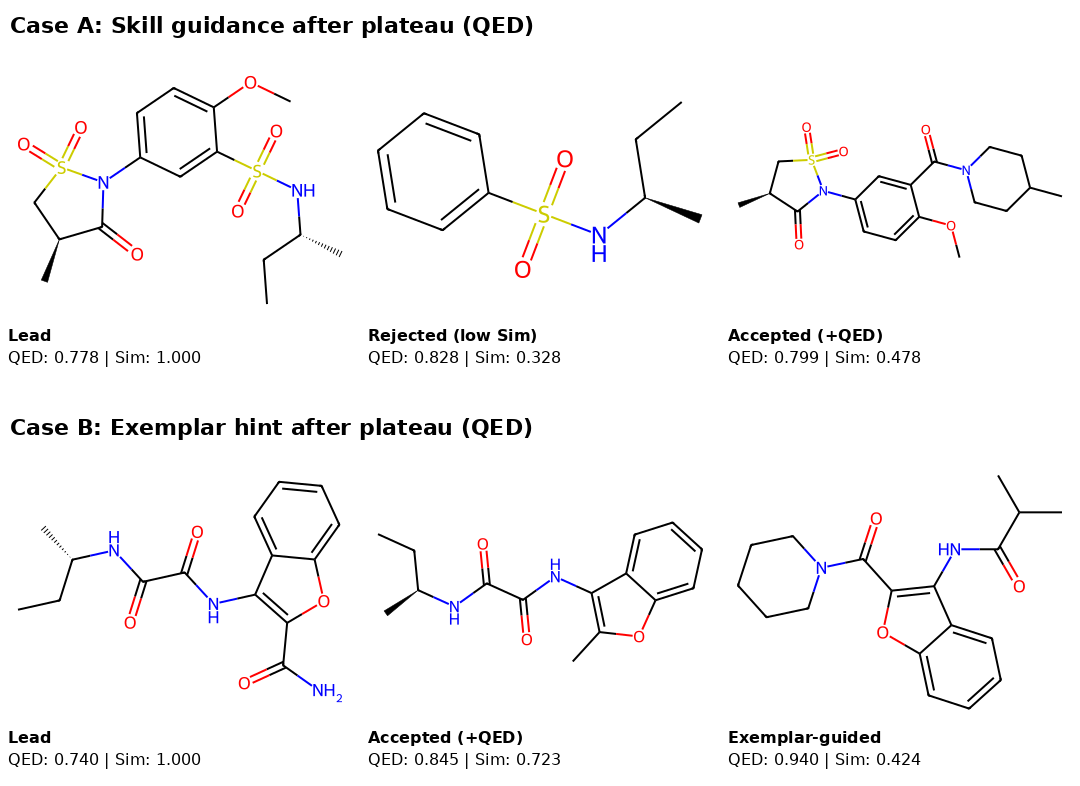}
\vspace{-3mm}
\caption{\small \textbf{Memory-triggered case studies.}
(Top) A QED run where an early proposal violates the similarity constraint; after the agent plateaus, retrieved guidance leads to a feasible edit that improves QED.
(Bottom) A QED run where exemplar hints provide a viable direction for subsequent edits. Each panel reports the lead molecule, representative intermediate proposals, and the resulting QED and similarity to the lead.}
\vspace{-4mm}
\end{figure*}